\documentclass{article} 
\usepackage[final]{colm2026_conference}
\usepackage{booktabs}
\usepackage{multirow}
\usepackage{xcolor}
\usepackage{booktabs}
\usepackage{makecell}
\usepackage{rotating}
\usepackage{colortbl}
\usepackage{adjustbox}
\usepackage{graphicx}   
\usepackage{subcaption}
\usepackage{longtable}
\usepackage{lscape}

\graphicspath{{plots_figures/}}

\definecolor{highrisk}{HTML}{F4CCCC}
\definecolor{modrisk}{HTML}{FCE5CD}
\definecolor{lowrisk}{HTML}{D9EAD3}
\usepackage{microtype}
\usepackage{hyperref}
\usepackage{url}
\usepackage{booktabs}

\usepackage{xcolor}
\usepackage{pifont}
\newcommand{\cont}{\textcolor{red}{\ding{55}}}
\newcommand{\notcont}{\textcolor{green!60!black}{\ding{51}}}

\usepackage{lineno}

\definecolor{darkblue}{rgb}{0, 0, 0.5}
\hypersetup{colorlinks=true, citecolor=darkblue, linkcolor=darkblue, urlcolor=darkblue}

\title{The State and Fate of Multilingual, Contextual Evaluation in the NLP World}

\author{Manan Uppadhyay\textsuperscript{$\heartsuit$}, Himanshu Beniwal\textsuperscript{$\heartsuit$$\clubsuit$}\thanks{Work done during internship at MSR India.}, Prashant Kodali\textsuperscript{$\heartsuit$}, Sunayana Sitaram\textsuperscript{$\heartsuit$}\\
\textsuperscript{$\heartsuit$}Microsoft Research India, \textsuperscript{$\clubsuit$}IIT Gandhinagar\\
\texttt{\{t-muppadhyay, b-hbeniwal, t-prakodali, sunayana.sitaram\}@microsoft.com} \\
}

\begin{document}

\ifcolmsubmission
\linenumbers
\fi

\maketitle

\begin{abstract}
Multilingual evaluation benchmarks are the primary instrument for assessing whether large language models generalize beyond English, yet the adequacy of these benchmarks has received little systematic scrutiny. We present a data-driven audit of 51 recent multilingual benchmarks spanning 242 datasets and 219 languages, organized around three pillars: \textit{coverage}, \textit{representativeness}, and \textit{rigor}. Our analysis reveals that coverage is wide but thin with 36\% of evaluated languages appearing in only a single benchmark, entire regions (Oceania, the Americas, Central Asia) are near-zero, and a stark task equity gap leaves low-resource languages evaluated on only 1--3 task categories versus 14 for high-resource languages. Representativeness is structurally compromised: translation from English remains the dominant construction strategy where 56\% of all dataset--language instances are translated introducing artifacts and English-centric framing, while culturally grounded content is concentrated in a handful of community-driven benchmarks with narrow language scope. The ecosystem thus forces a trade-off between breadth and validity. Rigor is undermined by benchmark contamination, including translated benchmark leakage and parallel corpus overlap that evade surface-form detection. We synthesize these findings into concrete recommendations for building evaluation frameworks that are natively constructed, culturally grounded, contamination-aware, and designed to serve the communities whose languages they claim to evaluate. The dashboard with all the metadata collected can be found at \href{https://aka.ms/multilingual-evals}{https://aka.ms/multilingual-evals.}
\end{abstract}

\section{Introduction}

Evaluation benchmarks are the primary instrument through which the NLP community measures progress. AI evaluation is critical beyond English, particularly for Global South languages, because most AI systems are designed from an English-centric perspective, with Global North assumptions baked into their data, task design, and notions of success \citep{akindotuni2025resource, lamentillo2025linguistic}. Benchmark gains on English can obscure systematic failures in other languages and conditions where AI systems are deployed, potentially leading to marginalized communities being excluded, stereotyped, or exposed to high-stakes failures \citep{lachini2024rethinking, ahmad2025bias}. Multilingual evaluation for Large Language Models (LLMs) has expanded rapidly in recent years, reflecting a growing recognition that English-only evaluation is insufficient \citep{weidinger2022taxonomy, mergen2025artificial, guo2025large, upadhayay2025efficient, kargaran2025mexa, zhang2026progressive}. However, multilingual evaluation is not simply about measuring in another language what is measured in English, but about capturing the local contexts and cultural nuances that determine quality, usefulness and safety. Recently at the India AI Impact Summit\footnote{\url{https://impact.indiaai.gov.in/}}, the New Delhi Frontier AI Commitments make this explicit through the commitment titled ``Strengthening Multilingual and Contextual Evaluations,'' which calls for ensuring AI system effectiveness ``across languages, cultures, and real-world use cases''\footnote{\url{https://www.pib.gov.in/PressReleasePage.aspx?PRID=2230201}}.

Progress on multilingual evaluation remains limited in important ways \citep{wu2025bitter, sinha2025eka, qin2025survey}, with many efforts still being Western-centric in their assumptions, priorities, and task formulations, often extending English benchmarks to other languages rather than grounding evaluation in the linguistic, cultural, and practical realities of diverse communities \citep{smith2025cultural, oh2025culture, mushtaq2025worldview, zahraei2025aligned, plum2025identity}. As a result, existing benchmarks are often not broad enough in the languages, varieties, and tasks they cover, nor sufficiently representative of how people actually use language technologies in different contexts \citep{mcintosh2025inadequacies, ni2025survey, yang2025socially}. In the spirit of \cite{joshi-etal-2020-state} "The State and Fate of Linguistic Diversity and Inclusion in the NLP World", which took stock of languages represented in NLP and ones that are left behind, we offer a similar reckoning for multilingual evaluation by examining its growth and gaps. We examine the current state of multilingual LLM evaluation through three key lenses: coverage, representativeness, and rigor, identify limitations in current practice and outline directions for building evaluation frameworks that are more meaningful, valid, and globally relevant. 
We will release all structured metadata annotations, the full analysis codebase, and an interactive web dashboard that will allow users to explore benchmark coverage, representativeness, and per-language statistics (Appendix~\ref{sec:appendix-dashboard}).

\section{Related Work}
\label{sec:related-work}
%


\paragraph{Multilingual Benchmarks and Language Coverage}
Early cross-lingual benchmarks such as XNLI \citep{conneau-etal-2018-xnli} established the paradigm of translating English datasets into multiple languages to perform multilingual evaluation. This was scaled across dozens of languages and task types by large benchmarking efforts such as XTREME-R, XGLUE, and MEGA \citep{ruder-etal-2021-xtreme, liang-etal-2020-xglue, ahuja-etal-2023-mega}. However, language coverage remains heavily skewed toward high-resource Indo-European and CJK languages \citep{joshi-etal-2020-state, blasi-etal-2022-systematic, conneau2020unsupervised, ruder-etal-2021-xtreme}, and low-resource languages continue to lag behind \citep{ahuja-etal-2023-mega, asai2024buffet, li2025s}. LLM-powered data augmentation for multilingual commonsense reasoning also fails to produce meaningful text in certain low-resource languages \citep{doostmohammadi2026toward, gain2025bridging}. Furthermore, multilingual LLMs have been shown to use English as an internal pivot, with abstract representations closer to English than to other input languages, revealing a structural English-centrism that needs to be addressed in evaluation \citep{wendler-etal-2024-llamas, ye2025multilinguality, blasi-etal-2022-systematic, kumar2025bridging, ahuja-etal-2023-mega}. These limitations have prompted calls for alternative paradigms such as performance prediction over per-language evaluation sets \citep{ahuja-etal-2022-beyond}.

\paragraph{Representativeness: Translation Artifacts and Cultural Grounding}
A critical concern with multilingual benchmarks is their reliance on translation from English \citep{huang2025tibetan}, as exemplified by XNLI's use of professionally translated MultiNLI data \citep{conneau-etal-2018-xnli}. TyDiQA \citep{clark-etal-2020-tydi} addressed this by eliciting questions directly in each target language, avoiding translationese\footnote{Translationese refers to unnatural linguistic patterns arising from translation that fail to reflect how native speakers naturally express themselves.} artifacts and better capturing language-specific information needs. Several studies further highlight how cultural knowledge, pragmatic norms, and world knowledge diverge from the English-centric assumptions embedded in translated datasets \citep{hershcovich-etal-2022-challenges, ruder-etal-2021-xtreme, conneau-etal-2018-xnli, hou2026bridging, ebrahimi2022americasnli, wendler-etal-2024-llamas}. Language-specific benchmarks such as MasakhaNER, IndicNLPSuite, AmericasNLI, and NusaX have shown that culturally grounded evaluation yields substantially different performance profiles than translated counterparts \citep{adelani-etal-2021-masakhaner, kakwani2020indicnlpsuite, ebrahimi2022americasnli, winata-etal-2023-nusax}. However, even these benchmarks do not always reflect the priorities or lived realities of the communities they target. Community-driven efforts like Pariksha \citep{watts-etal-2024-pariksha} and Samiksha \citep{bhat2025building} help bridge this gap by incorporating stakeholder input, though such benchmarks remain scarce. More recently, benchmarks like TUMLU \citep{isbarov-etal-2025-tumlu} for Turkic languages have begun prioritizing sociolinguistic authenticity over mere cross-lingual alignment.

\paragraph{Evaluation Rigor and Data Contamination}

The integrity of benchmark-based evaluation has come under increasing scrutiny as LLMs are trained on ever-larger web corpora that may include benchmark test data. Contamination has been argued to represent an existential threat to LLM evaluation, necessitating systematic analysis before conclusions are drawn from benchmark results \citep{ahuja2024contaminationreportmultilingualbenchmarks, sainz-etal-2023-nlp, jacovi-etal-2023-stop, ahuja-etal-2023-mega}. Practical mitigation strategies have been proposed, including encrypting test data and demanding training exclusion controls from closed API model builders to proactively protect test data rather than retroactively detect leakage \citep{jacovi-etal-2023-stop, sainz-etal-2023-nlp, blasi-etal-2022-systematic}. The problem is further complicated by the opacity of training data in proprietary models, where contamination can only be inferred indirectly \citep{sainz-etal-2023-nlp, jacovi-etal-2023-stop, osti_2585515} and agentic behavior showing evidence of knowledge of being tested\footnote{\url{https://www.nist.gov/caisi/cheating-ai-agent-evaluations}} and deliberate cheating\footnote{\url{https://www.anthropic.com/engineering/eval-awareness-browsecomp}}. In the multilingual setting, contamination risks are compounded by the existence of parallel corpora and translated benchmark variants: a model trained on English test data may exhibit inflated performance on translated versions in other languages, yet this cross-lingual leakage pathway remains poorly characterized \citep{sainz-etal-2023-nlp, liang-etal-2020-xglue, ruder-etal-2021-xtreme, conneau-etal-2018-xnli, cruz2026llm}. Contamination detection methods trained on one language often fail to generalize to linguistically dissimilar languages \citep{macko2023multitude} and usually rely on surface-form overlap, which is inadequate for detecting semantic leakage through paraphrase or translation \citep{cheng2025survey, macko2023multitude, blevins-zettlemoyer-2022-language, fu2025does, cruz2026llm}.


\section{Survey of Existing Multilingual Benchmarks}

We conduct a unified audit of existing multilingual benchmarks across coverage, representativeness, and rigor in four stages: benchmark collection, structured field extraction, manual verification, and contamination checks.

We compile a corpus of 51 publicly released multilingual evaluation benchmarks, spanning diverse tasks (NLI, QA, NER, sentiment analysis, commonsense reasoning, summarization, and others), language resource levels, and language families. We identify candidate benchmarks through a combination of literature search, leaderboard tracking, and community repositories. We include both widely cited benchmarks (e.g., XNLI, XQuAD, TyDiQA) and recent community-driven efforts targeting underrepresented languages and regions (e.g., AfriSenti, MasakhaNER, IrokoBench). The full list is provided in Table~\ref{tab:appendix-benchmarks}. For fine-grained analysis, each benchmark must be annotated with a consistent set of structured fields, including languages covered, language families, scripts, task types, construction methodology (translated vs.\ natively authored), data sources, and licensing information. To extract these fields at scale, we leverage Claude Opus 4.6\footnote{\url{https://www.anthropic.com/news/claude-opus-4-6}}, prompting the model with the benchmark's paper and metadata as context. The extraction methodology is provided in the Appendix~\ref{sec:appendix-prompt}. This semi-automated approach enables consistent, structured annotation across all 51 benchmarks while substantially reducing manual effort. We then perform a thorough manual verification of the annotations by cross-referencing the extracted fields with the original papers, dataset cards, and associated repositories.



\subsection{Coverage}
\label{sec:coverage}

We analyze 51 benchmarks encompassing 242 distinct datasets and 219 unique languages along four dimensions: language, typological and geographic diversity, task coverage, and resource-level balance. Because many benchmarks aggregate existing datasets (22 of the 51 are mixed or aggregation suites), we conduct our analysis at the \textit{dataset--language} level to avoid inflating coverage counts when the same dataset appears in multiple benchmarks.Unless otherwise specified, all subsequent analyses are conducted and reported at the dataset level. 

\begin{figure}[t]
    \centering
    \includegraphics[width=0.9\textwidth]{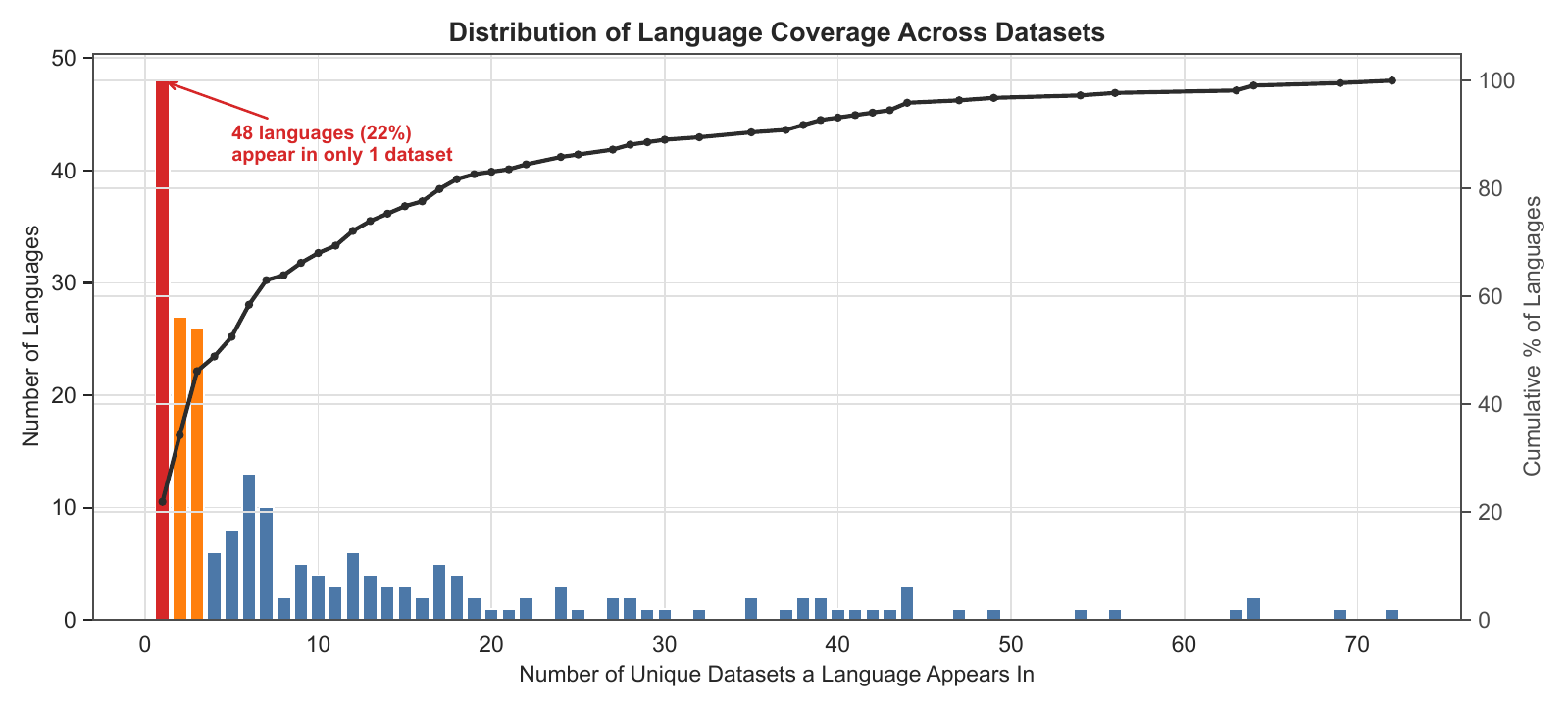}
    \caption{Languages ranked by the number of datasets in which they appear. \textit{The long tail reveals that 36\% of evaluated languages appear in only a single dataset}.}
    \label{fig:long-tail}
\end{figure}

\textbf{Language Coverage:}\label{sec:language-coverage} Across the 51 benchmarks we survey, a total of \textbf{219 unique languages} are evaluated across 242 datasets, suggesting reasonable coverage. However, the distribution follows a steep power law, as shown in Figure~\ref{fig:long-tail} that plots languages ranked by the number of datasets in which they appear. A small cluster of high-resource languages Hindi, Chinese, Spanish, Swahili, Arabic appear in 19 or more datasets, while \textbf{48 of 219 languages (37\%) appear in only a single dataset}. For these 48 languages, there exists no independent replication of evaluation results and no opportunity for cross-dataset comparison. This long-tail structure has a direct methodological implication: claims of multilingual generalization rest on a dense evaluation of a handful of languages and single-point estimates for the majority. 


\begin{figure}[t]
    \centering
    \begin{subfigure}[t]{0.49\textwidth}
        \centering
        \includegraphics[width=\textwidth]{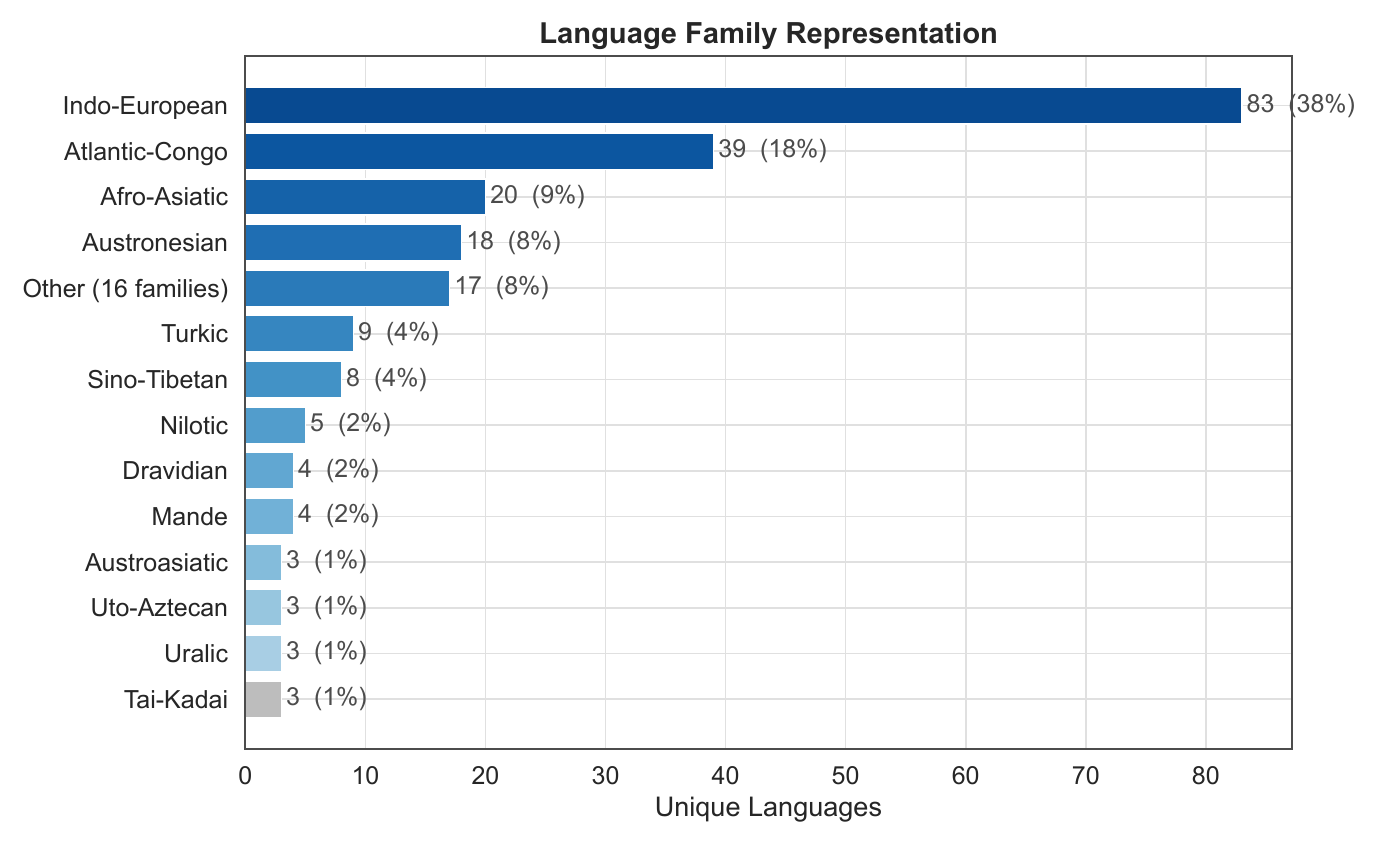}
        \caption{Language family representation.}
        \label{fig:family-coverage}
    \end{subfigure}
    \hfill
    \begin{subfigure}[t]{0.49\textwidth}
        \centering
        \includegraphics[width=\textwidth]{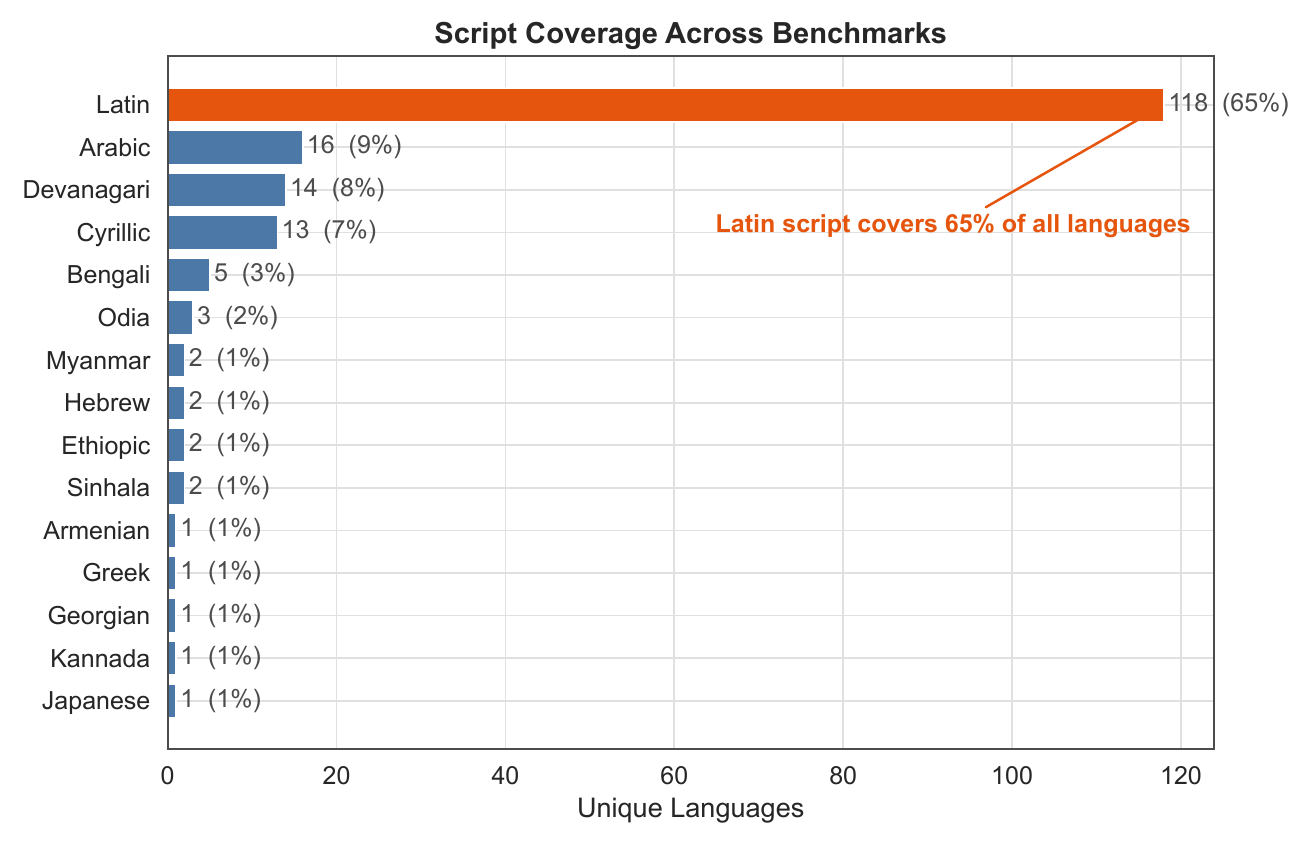}
        \caption{Script coverage (top 15 scripts).}
        \label{fig:script-coverage}
    \end{subfigure}
    \caption{Typological concentration in multilingual benchmarks. \textbf{(a)} \textit{Indo-European dominates at 38\% of all languages, with many families represented by fewer than 5 languages.} \textbf{(b)} \textit{Latin script accounts for the majority; widely-used scripts such as Ge'ez, Khmer, and Myanmar are barely represented.}}
    \label{fig:typological-coverage}
\end{figure}

\textbf{Typological and Geographical Coverage:} \label{sec:typological-coverage} Language count alone is an insufficient measure of diversity. Two benchmark suites could each cover 20 languages yet differ vastly in typological range if one samples broadly across families and scripts while the other concentrates on closely related Indo-European varieties. Figure~\ref{fig:family-coverage} shows the distribution of unique languages across the 29 language families represented in our survey. Indo-European languages account for \textbf{83 of 219 languages (38\%)}, while Atlantic-Congo (39 languages) ranks second, but this representation is almost entirely attributable to Africa-focused benchmarks like AfroBench, IrokoBench, and Uhuru. Families such as Austroasiatic (3 languages), Tai-Kadai (3), Japonic (1), and Quechuan (2) are poorly represented. The entire Oceanic branch of Austronesian is absent. Figure~\ref{fig:script-coverage} reveals a similar distribution: Latin script accounts for the majority of evaluated languages, followed by Devanagari, Arabic, and CJK. Scripts used by hundreds of millions of speakers such as Ge'ez (Ethiopic), Khmer, Myanmar appear in only 1--3 benchmarks. Script diversity matters because models may exhibit systematically different tokenization efficiency~\citep{ahia-etal-2023-languages}, character-level recall, and decoding accuracy across scripts, and these script-specific failure modes remain invisible when evaluation concentrates on Latin and Devanagari. We find similar trends in Geographical coverage as well with entire regions of the world such as Oceania and South America having very poor or no representation.

\begin{figure}[t]
    \centering
    \includegraphics[width=0.9\textwidth]{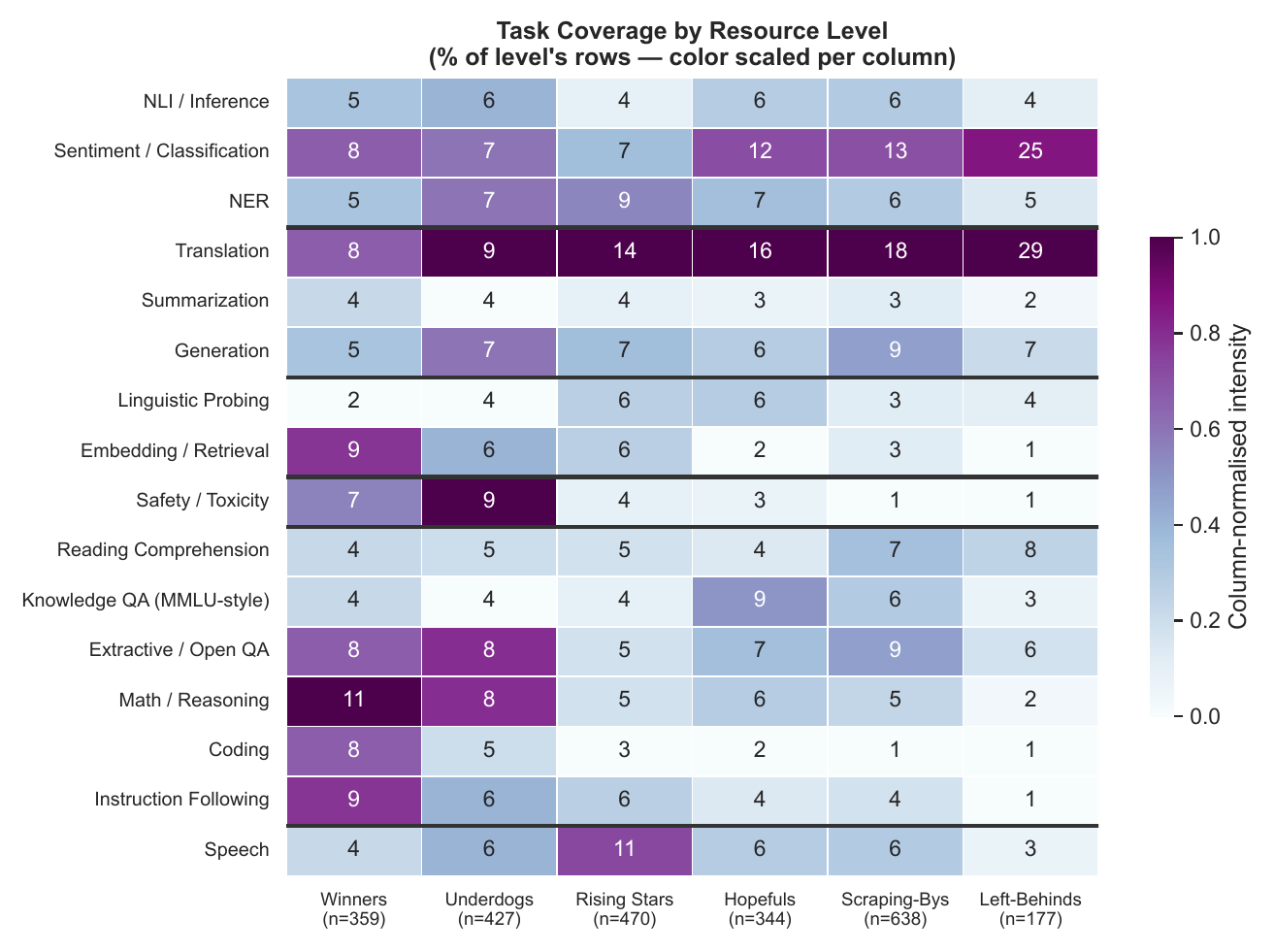}
    \caption{Task coverage by resource level. Each cell shows the percentage of a resource level's dataset--language rows devoted to that task category. \textit{Left-Behinds are overwhelmingly evaluated on Translation (29\%) and Sentiment/Classification (25\%); specialised tasks such as Coding, Safety, Instruction Following, each account for $\leq$1\% of their evaluation, whereas Winners are evaluated more evenly across all 15 categories.}}
    \label{fig:task-equity-fine}
\end{figure}

\textbf{Resource-Level Distribution and Task Equity:}\label{sec:resource-level}\label{sec:task-coverage} We examine the resource-level composition of evaluated languages using the taxonomy from~\citet{joshi-etal-2020-state}. Across the 242 distinct datasets, 65\% of evaluated languages fall in the two lowest resource classes (Scraping-Bys and Left-Behinds, levels 0--1), while Winners (level 5) account for only 5\% (Figure~\ref{fig:resource-pie-appendix} in Appendix~\ref{sec:appendix-geographic}). This distribution appears to favor low-resource languages, but a finer-grained analysis reverses the picture. Figure~\ref{fig:task-equity-fine} disaggregates task coverage across 15 categories by resource level: the 10 Winner languages are each evaluated on a median of 14 task categories across multiple benchmarks, whereas the 160 low-resource languages (levels 0--2) are typically tested on only 1--3 tasks in a single benchmark. For Left-Behinds, Translation alone accounts for 29\% of all evaluation rows and Sentiment/Classification for 25\%, while Coding, Safety/Toxicity, and Instruction Following each represent only 1\%. By contrast, Winners and Underdogs distribute evaluation more evenly, with Math/Reasoning (11\% and 8\%), Embedding/Retrieval (9\% and 6\%), and Safety (7\% and 9\%) all receiving meaningful attention. In absolute terms, Safety/Toxicity evaluation is scarce---only roughly 25 dataset--language rows across all datasets---and these are concentrated almost entirely at higher resource levels, meaning that safety coverage for low-resource languages is effectively nonexistent. Figure~\ref{fig:task-region-appendix} (Appendix) cross-tabulates task categories against world regions and confirms that QA and knowledge tasks dominate everywhere, while specialized capabilities show uneven coverage. This \textit{task equity gap} means that for most low-resource languages, a model's ``evaluation'' amounts to a narrow probe on translation and classification, making it far too limited to support claims of general competence. 


The coverage analysis reveals an ecosystem that is \textit{wide but thin}: many languages are nominally included, but the depth of evaluation measured by replication across datasets, typological diversity, task breadth, and resource-level balance is concentrated in a small core. The next section asks whether the quality of coverage that does exist is adequate.

\subsection{Representativeness}
\label{sec:representativeness}

Coverage establishes \textit{what} is evaluated; representativeness asks whether that evaluation is \textit{valid}. A dataset may nominally cover a language, yet evaluate it using content translated from English, written by non-native speakers, or detached from local cultural context and assumptions. We audit the 242 datasets along two axes: translation and cultural grounding. At the dataset--language level, 56\% of all instances are translated from English, 36\% are natively authored, and 8\% are partially adapted.

\begin{figure}[t]
    \centering
    \begin{subfigure}[t]{0.48\textwidth}
        \centering
        \includegraphics[width=\textwidth]{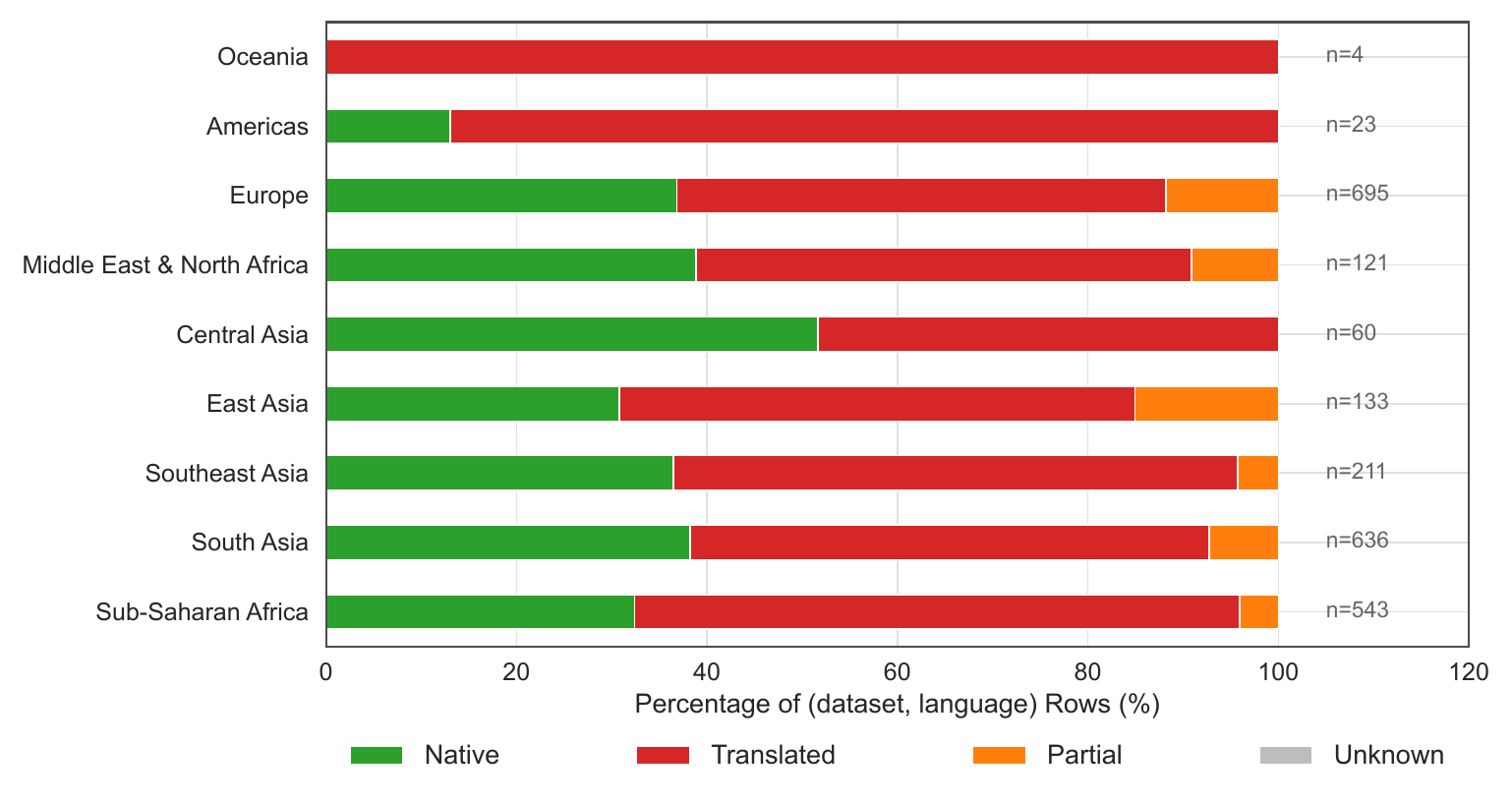}
        \caption{Translation status by world region.}
        \label{fig:translation-region}
    \end{subfigure}
    \hfill
    \begin{subfigure}[t]{0.48\textwidth}
        \centering
        \includegraphics[width=\textwidth]{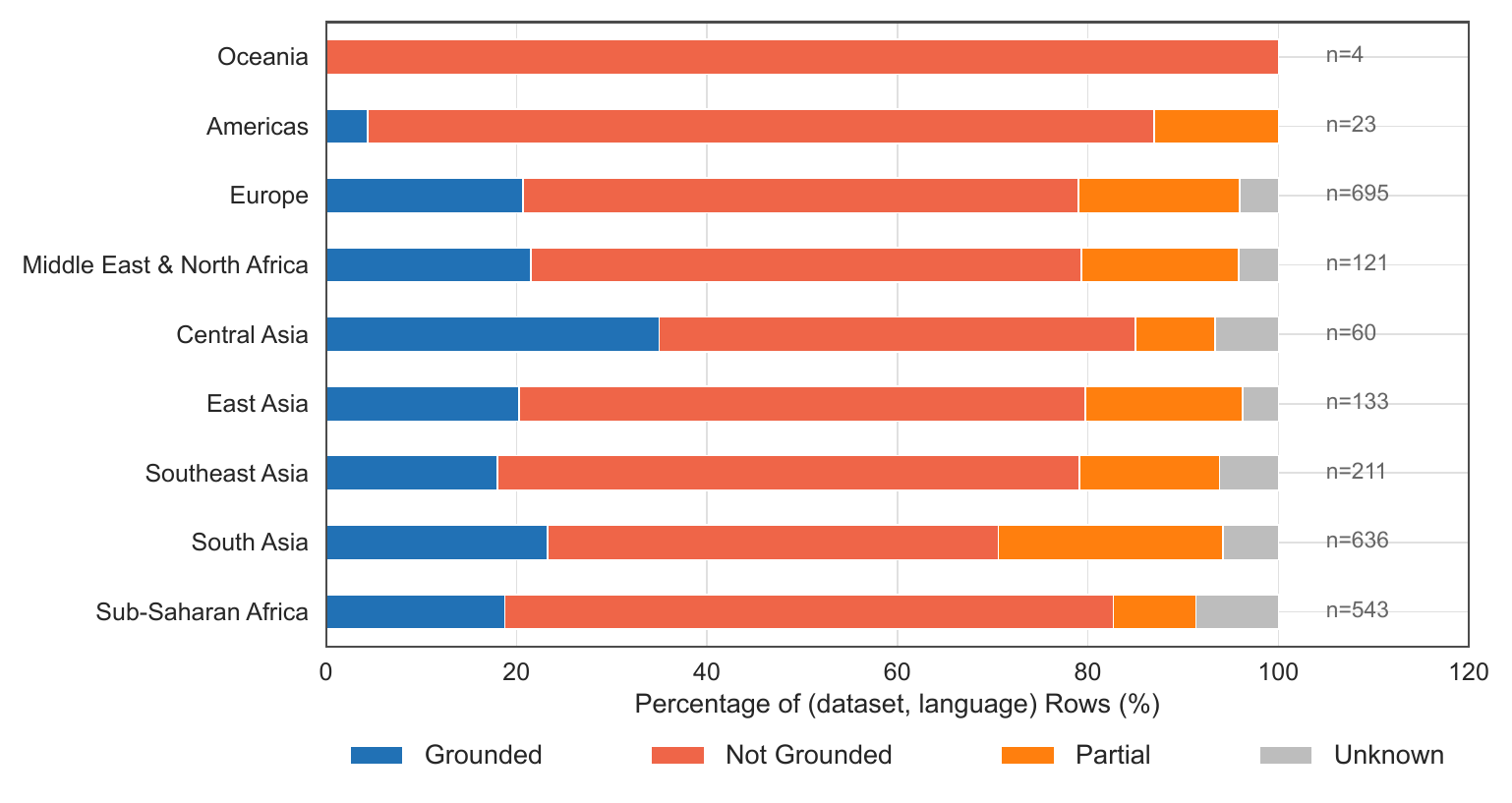}
        \caption{Cultural grounding by world region.}
        \label{fig:cultural-region}
    \end{subfigure}
    \caption{Representativeness of multilingual benchmarks by world region. \textbf{(a)} \textit{Europe and East Asia are the most translation-dependent; Sub-Saharan Africa has the highest share of natively authored content.} \textbf{(b)} \textit{Community-driven benchmarks in Sub-Saharan Africa and South Asia achieve the highest cultural grounding rates; broad-coverage suites score lower.}}
    \label{fig:representativeness-region}
\end{figure}

\textbf{Translation vs.\ Native Construction:} \label{sec:translation} The dominant construction strategy for multilingual datasets remains translation from English source datasets. This approach scales efficiently as one annotation effort produces content in many languages but comes with the cost of translationese artifacts, English-centric framing, and inherited contamination risk (Section~\ref{sec:rigor})~\citep{artetxe-etal-2020-translation, graham-etal-2020-statistical}. Figure~\ref{fig:translation-region} disaggregates translation status by world region. In every region, a substantial fraction of languages are evaluated primarily through translated content. Datasets covering Europe and East Asia are the most translation-dependent, while Sub-Saharan Africa shows the highest share of natively authored content, which is a direct consequence of language-specific datasets created from scratch such as AfroBench~\citep{ojo-etal-2025-afrobench}, IrokoBench~\cite{adelani-etal-2025-irokobench}, and MasakhaNER~\citep{adelani-etal-2021-masakhaner}.


\textbf{Cultural Representativeness: } \label{sec:cultural} Culturally grounded datasets reflect the norms, knowledge, and pragmatic conventions of the target language community. A dataset can be natively authored yet culturally shallow (e.g., factoid QA drawn from Wikipedia), and a translated dataset can incorporate cultural adaptation (transcreation) to make it more culturally grounded. However, in practice, the two are correlated: datasets constructed from scratch by native-speaker communities are far more likely to be culturally grounded. Figure~\ref{fig:cultural-region} shows cultural grounding rates by region. Datasets targeting Sub-Saharan Africa and South Asia exhibit the highest cultural grounding rates, reflecting the deliberate design philosophy of community-driven efforts like CulturalBench~\citep{chiu-etal-2025-culturalbench}, Samiksha ~\cite{hamna2026buildingbenchmarksgroundup}, and TUMLU~\citep{isbarov-etal-2025-tumlu}. By contrast, datasets with broad cross-regional scope tend to have lower cultural grounding as it is difficult to achieve at scale. This finding reinforces the pattern: the datasets that are methodologically strongest along the representativeness axis are those with the narrowest language coverage.

\textbf{Annotator Demographics:} Transparency around annotation workforces remains limited: only 8 of 51 benchmarks (16\%) report annotator demographics such as geographic origin, ethnicity, or educational background, and only 17 (33\%) disclose compensation details. Researchers are the dominant creator type across datasets, with community volunteers and crowdworkers accounting for a small minority, raising questions about whose linguistic backgrounds and priorities shape ``ground truth'' for underrepresented languages. The few benchmarks that do document their annotation pipeline in detail (CulturalBench, Aya, and Uhuru) tend to be the same ones that score highest on cultural grounding, suggesting that annotator transparency and evaluation validity go hand in hand.

\subsection{Rigor}
\label{sec:rigor}

\subsubsection{Metrics}

Metrics determine what counts as success, and poorly chosen metrics can give a misleading picture of model quality. Most benchmarks rely either on standard reference-based metrics to compare model outputs against ground truth, or on reference-free metrics such as LLM judges. However, many widely used metrics, such as BLEU, ROUGE, exact match, and Word Error Rate, were developed largely in English-centric settings and can be poor proxies for quality in morphologically rich languages or languages with non-standard spellings, where valid responses may differ substantially in surface form. Similarly, although LLM judges are increasingly becoming standard in evaluation pipelines, prior work shows that they often align poorly with human judgments in multilingual and multicultural settings, particularly for low-resource languages and in culturally grounded, context-sensitive evaluations \citep{hada-etal-2024-large, watts-etal-2024-pariksha, hamna2026buildingbenchmarksgroundup, hada-etal-2024-metal}.

\subsubsection{Contamination}

\begin{table}[t]
\centering

\small
\begin{tabular}{llccc}
\toprule
	\textbf{Dataset} & \textbf{Task} & \textbf{Prior Work} & \textbf{STAMP} & \textbf{DCQ} \\
\midrule
	\textsc{Paws-X}      & Paraphrase Det. & 9/10  & 67.9\% & 32.6\% \\
	\textsc{Mgsm}        & Math            & ---   & 45.5\% & 30.9\% \\
	\textsc{XCopa}       & Commonsense     & 9/10  & ---    & ---    \\
	\textsc{XQuAD}       & QA              & 9/10  & ---    & ---    \\
	\textsc{AfriMGSM}    & Math            & ---   & 21.0\% & 18.4\% \\
	\textsc{Xnli}        & NLI             & 7/10  & ---    & ---    \\
	\textsc{XStoryCloze} & Story Cloze     & 7/10  & ---    & ---    \\
	\textsc{Flores}      & Translation     & 7/7   & 1.2\%  & ---    \\
	\textsc{XLSum}       & Summarization   & 5/7   & ---    & ---    \\
\midrule
	\textsc{IndicParam}  & Knowledge       & ---   & 4.3\%  & 14.7\% \\
	\textsc{Mmmlu}       & Knowledge       & ---   & 2.0\%  & 19.6\% \\
	\textsc{Global MMLU} & Knowledge       & ---   & 1.7\%  & 17.5\% \\
	\textsc{Include}     & Knowledge       & ---   & 1.0\%  & 18.8\% \\
	\textsc{Milu}        & Knowledge       & ---   & 0.0\%  & 21.7\% \\
\bottomrule
\end{tabular}
\caption{Summary of contamination evidence across 14 multilingual datasets. \textbf{Prior Work}: fraction of models flagged as contaminated by surface-form and overlap methods~\citep{ahuja-etal-2024-megaverse, ahuja2024contaminationreportmultilingualbenchmarks, yao-etal-2024-data} (details in Table~\ref{tab:contamination_combined}). \textbf{STAMP}: overall contamination rate (\% of model--language pairs flagged at $p{<}0.05$) from our analysis using \textsc{STAMP}~\citep{rastogi2025stamp} (details in Table~\ref{tab:contamination_results}). \textbf{DCQ}: mean maximum recognition rate across models and languages using the Data Contamination Quiz~\cite{}; 20\% is the random-chance floor (details in Table~\ref{tab:dcq_results}).}
\label{tab:contamination_summary}
\end{table}

Contamination is especially consequential in multilingual evaluation because much of the evaluation ecosystem is \textit{thin}: \textbf{36\% of languages} appear in only one dataset (Figure~\ref{fig:long-tail}), and \textbf{56\% of dataset--language rows} rely on translated content (Figure~\ref{fig:translation-region}). When a language's evaluation rests on a single translated dataset, contamination through any pathway \textbf{invalidates the only evidence of model competence} for that language. The \textbf{task equity gap} (Figure~\ref{fig:task-equity-fine}) further amplifies this risk: for low-resource languages evaluated on only 1--2 task categories, contamination on even a single task can dominate the reported result.

{\bfseries Prior Contamination Analyses.} Several studies have investigated multilingual benchmark contamination using surface-form overlap and membership inference methods. \citet{ahuja-etal-2024-megaverse} found that GPT-4, Llama-2, and Gemini-Pro are likely contaminated on PAWS-X, XCopa, and XQuAD. \citet{ahuja2024contaminationreportmultilingualbenchmarks} extended this analysis to newer models (Llama-3.1, Mistral-v0.3, Gemma-2) and confirmed pervasive contamination across 7 benchmarks. \citet{yao-etal-2024-data} further demonstrated that contamination \textit{transfers cross-lingually}: models contaminated on the English version of a dataset show inflated scores on non-English translations (Table~\ref{tab:contamination_crosslingual}, Appendix). Aggregating across these studies, 9 of the 10 models tested are contaminated on PAWS-X, XCopa, and XQuAD, and 7 of 7 are contaminated on Flores (Table~\ref{tab:contamination_summary}; per-model details in Table~\ref{tab:contamination_combined}, Appendix). However, these analyses examined a limited set of established datasets, leaving open the question of whether contamination extends to newer evaluation suites.

{\bfseries Contamination Analysis with STAMP.} To provide an updated assessment, we apply \textbf{STAMP}~\citep{rastogi2025stamp} to a broader and more contemporary suite of \textbf{9 datasets} including recent ones such as Global MMLU, MILU, INCLUDE, and IndicParam alongside established ones like PAWS-X, MGSM, and AfriMGSM across \textbf{7 recent models} (Table~\ref{tab:contamination_results}, Appendix). STAMP makes minimal assumptions about data format, enabling reliable membership detection at trace levels while strictly preserving the semantic meaning and evaluation utility of the original content. Following \citet{rastogi2025stamp}, we operationalize contamination detection by generating 9 paraphrased versions of each dataset item and measuring the performance gap between the original and its variants. Paraphrases are produced using \textbf{Qwen-3.5 397B}, a large multilingual model chosen because its scale ensures high-quality rephrasings across diverse writing systems. Our results reveal that contamination remains \textbf{pervasive} for older datasets: \textbf{PAWS-X} (67.9\% of model--language pairs flagged), \textbf{MGSM} (45.5\%), and \textbf{AfriMGSM} (21.0\%) show significant contamination signals across all seven models (Table~\ref{tab:contamination_summary}). By contrast, newer datasets---MILU (0.0\%), INCLUDE (1.0\%), and Global MMLU (1.7\%) exhibit markedly lower contamination rates, suggesting that dataset age and prominence are strong predictors of contamination risk. One notable discrepancy emerges for \textbf{Flores}: prior surface-form methods flag 7 of 7 models, yet STAMP detects contamination in only 1.2\% of model--language pairs. This divergence likely reflects differences in what each method measures, i.e. surface-form overlap versus memorization-driven performance gaps and underscores that no single detection approach suffices for multilingual contamination.

{\bfseries Complementary Analysis with DCQ.} We additionally apply the \textbf{Data Contamination Quiz (DCQ)}~\cite{golchin-surdeanu-2025-data} to 85 dataset--language partitions from 8 datasets across the same 7 models. For each of 100 sampled items per partition, GPT-4.1 produces four word-level synonym perturbations. A bias-detection quiz first presents only the four perturbations and a ``none of these'' option; the original item is then inserted into positions that the model under-selected in this initial quiz. Selecting the verbatim original provides a recognition signal while the initial responses estimate positional bias. DCQ reports a conservative and a maximum recognition rate; we use the maximum rate in Table~\ref{tab:contamination_summary} for consistency across models, while noting that taking the best rate across positions is optimistic and that 20\% is the random-chance floor. At the dataset level, PAWS-X (32.6\%) and MGSM (30.9\%) show the strongest average recognition, whereas the remaining dataset averages range from 14.7\% to 21.7\%. Recognition is generally higher for English partitions: for example, English versus non-English averages are 45.7\% versus 15.3\% for AfriMGSM, 44.0\% versus 29.6\% for MGSM, and 47.6\% versus 30.1\% for PAWS-X. These values measure recognition rather than the fraction of a dataset that leaked and should not be directly equated with STAMP's significance-based flag rate. Full dataset--model results, language-group comparisons, and validity caveats are reported in Appendix~\ref{sec:appendix-dcq}.

{\bfseries Challenges of Multilingual Contamination Detection.} Both STAMP and DCQ rely on another LLM to construct the measuring instrument, so variation in that model's competence across languages can be mistaken for variation in contamination. For STAMP, semantically faithful watermarked rephrasing is harder across typologically distant and low-resource languages; uneven paraphrase quality introduces noise into the paired test. Its KGW watermarking scheme also partitions vocabulary through a tokenizer-dependent hash function~\citep{pmlr-v202-kirchenbauer23a}, so sparse or unstable tokenization can weaken the watermark and the test's statistical power. For DCQ, difficulty is determined by the quality of GPT-4.1's decoys. Perturbation distance ranges from 5.8\% for AfriMGSM in Amharic to 48.1\% for Global MMLU in English and varies with language resource level. In the eight weakest partitions, decoys are too similar to the original for reliable separation, pinning recognition near the chance floor regardless of exposure; conversely, misspellings or semantic drift can make decoys easier to reject and inflate recognition. This asymmetry is especially consequential because it can reproduce the expected finding: apparently lower contamination in low-resource languages may reflect a weaker measuring instrument rather than less exposure.

DCQ introduces a further identification problem because its prompt names the dataset, potentially conflating recognition of the dataset label with recognition of its instances. Our natural replicate holds all 100 GSM8K items byte-identical while changing only the label from AfriMGSM to MGSM; the resulting recognition difference ranges from 1 percentage point for Aya-Expanse-8B to 23 points for Gemma-3-12B (Appendix~\ref{sec:appendix-dcq}). This variation cannot be attributed to contamination and limits how finely models or languages can be compared. Removing the dataset name would reduce the confound but also remove a cue on which DCQ's recognition design depends. Together, these limitations show that neither method is language-neutral: multilingual contamination claims require per-language instrument validation and convergence across complementary methods rather than direct comparison of raw scores.

\subsubsection{Reporting Practices in Model Releases}
\label{subsec:reporting-practices}

Based on our analysis, we ask: how do model developers report multilingual evaluation? We audit 23 recent model releases (15 open-weight, 8 closed-models, 2024--2026) across five binary transparency dimensions (1) whether training language composition is explicitly disclosed, (2) whether any multilingual evaluation dataset/benchmark is reported, (3) whether the specific evaluation languages are listed, (4) whether a contamination detection technique is named, and (5) whether contamination results are reported. The composite transparency score is the sum (0–5) (Figure~\ref{fig:reporting-transparency}; full methodology and per-model breakdown in Appendix~\ref{sec:appendix-reporting}). The results are stark: 35\% of releases report \textit{zero} multilingual datasets, and 87\% provide no contamination analysis. Only 3 models (all open-weight) report any contamination testing, and even these leave cross-lingual contamination pathways unexamined. When multilingual evaluation \textit{is} reported, it concentrates on a handful of benchmarks, the model--dataset matrix is 93\% sparse, while natively authored datasets from our survey (AfroBench, IrokoBench, MILU) are almost entirely unused. Open-weight models are more transparent (mean score 2.47/5) than closed (0.75/5), but both fall well below adequacy: the overall mean is 1.87/5, and no closed model discloses training language composition or reports contamination. These findings argue for a standardized multilingual evaluation card mandating per-language dataset disclosure accounting for contamination. Finally, multilingual safety evaluation is reported by only 4 of 23 models: Tiny~Aya, Claude~Opus~4.6, Claude~Sonnet~4.6, and CommandA. All other models either test safety only in English or do not report safety testing at all.

\section{Discussion}
\label{sec:discussion}

Our audit of 51 recent multilingual benchmarks spanning 242 datasets reveals a consistent pattern across all three pillars. \textbf{Coverage} is wide but thin: 36\% of languages appear in only a single dataset and entire regions of the world are completely absent. Multilingual safety datasets remain few in number and provide little coverage of low-resource languages. \textbf{Representativeness} is structurally compromised: the datasets with the broadest language coverage are the most translated and least culturally grounded, while the most valid datasets cover the fewest languages. Annotator demographics are rarely reported, with researchers being the primary data creators of multilingual datasets, raising concerns about whose values and priorities are reflected in these datasets. \textbf{Rigor} is undermined by multiple contamination pathways that are amplified by translation-based construction, English-centric metrics that do not always work well on diverse languages and a lack of meaningful reporting of multilingual evaluation in model releases. These findings point to a systemic misalignment between how multilingual evaluation is currently conducted and what it would need to be to support trustworthy claims of cross-lingual competence. If these issues are not taken seriously, multilingual evaluation risks becoming performative: creating the appearance of inclusion while masking shallow support, overstating progress, and ultimately widening existing linguistic and cultural gaps in AI.

Supporting a language is not simply a matter of translating prompts, reporting a score, or listing it in a benchmark; it requires evidence that the system works in ways that are meaningful for users in their local linguistic and cultural contexts. Similarly, evaluating a language is not just measuring in another language what was originally defined in English, but assessing model behavior under the norms, knowledge systems, communicative practices, and risks that shape real use. Our work aims to raise the bar for how multilingual evaluation is conducted and reported. Reporting should make clear what was evaluated, how data was created or adapted, what contexts are covered, and what claims the results do and do not support. Meaningful community participation must go beyond translation or validation to include communities in defining tasks, rubrics, and failure modes. AI itself can also help scale multilingual evaluation through advances in multilingual synthetic data \citep{chitale2026updeshsynthesizinggroundedinstruction}, aligned multilingual LLM judges, and dynamic benchmarks to avoid contamination. Multilingual evaluation must also move beyond model-level assessment to include product, user, and real-world impact evaluation\footnote{\url{https://eval.playbook.org.ai/}}, since model performance alone cannot capture how systems behave in deployed, culturally situated settings.

There is an inherent tension between coverage and representativeness in multilingual evaluation: approaches that scale across many languages often depend on standardization, while representativeness requires deeper engagement with local variation, context, and heterogeneity. If the field is committed to supporting everyone with AI, it must accept this tension as a fundamental reality: meaningful evaluation cannot always be reduced to uniform or fully replicable patterns across highly diverse settings, and some degree of variation must be expected as a consequence of taking linguistic, cultural, and contextual diversity seriously.

\bibliography{colm2026_conference,anthology-1,anthology-2}
\bibliographystyle{colm2026_conference}

\appendix
\section{Appendix}
\subsection{Data Extraction Methodology}
\label{sec:appendix-prompt}

\begin{figure}[t]
    \centering
    \includegraphics[width=0.9\textwidth]{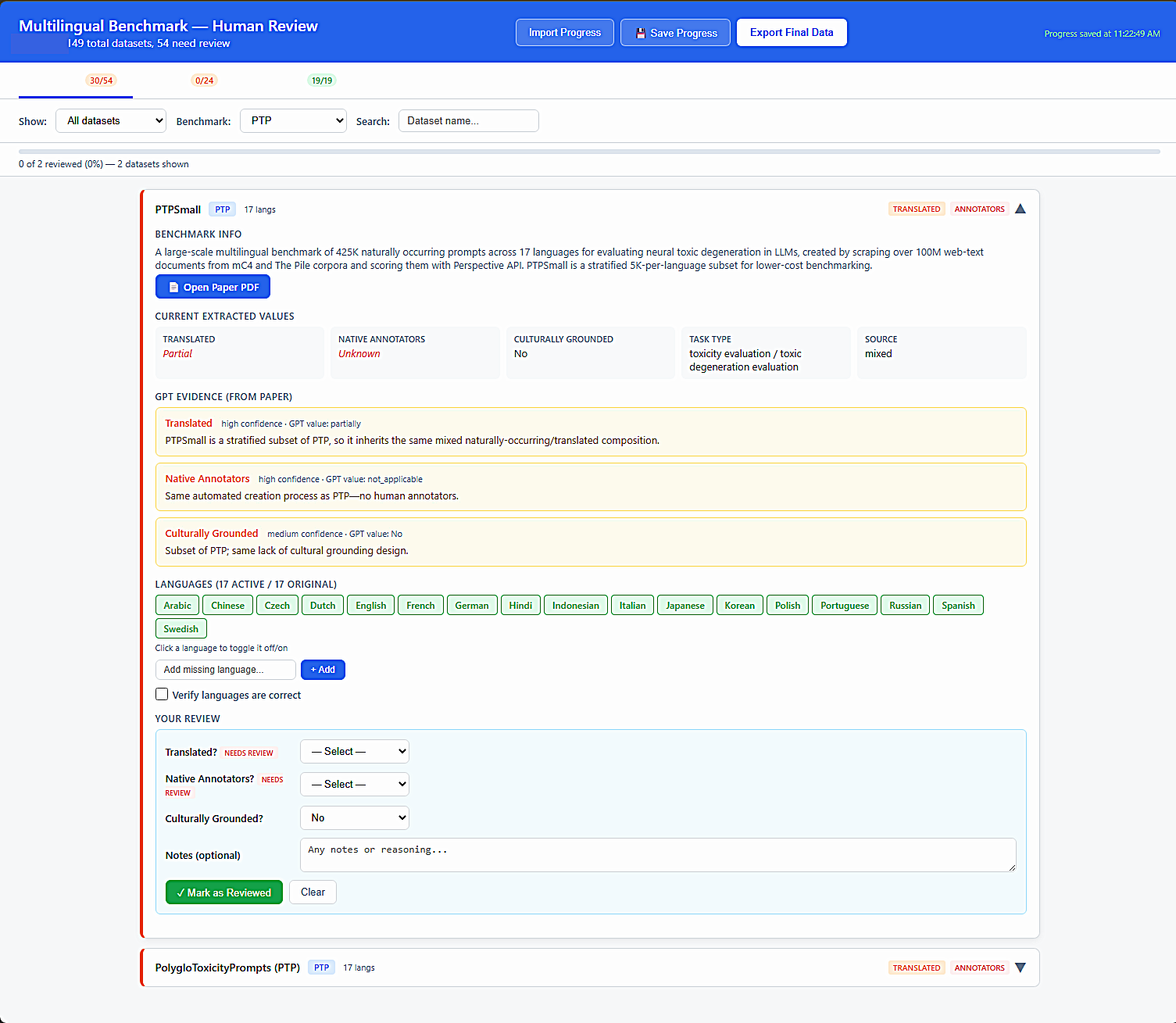}
    \caption{Human review interface showing extracted dataset properties with evidence passages and confidence ratings, alongside structured dropdowns for reviewer adjudication of translation status, native annotator involvement, and cultural grounding.}
    \label{fig:annotate-users}
\end{figure}

We survey 51 multilingual evaluation benchmarks and extract structured metadata at dataset-level granularity through a multi-step pipeline. For each benchmark/dataset paper, Claude Opus 4.6 extracts the constituent datasets, their member languages, and three key properties per dataset: whether the evaluation content was translated from another language (typically English), whether native speakers were involved in annotation, and whether the dataset is culturally grounded in the target locale rather than reflecting source-culture assumptions. The extraction operates at a three-level hierarchy—benchmark $\rightarrow$ dataset $\rightarrow$ language. Language metadata is enriched deterministically from authoritative sources: family and macroarea, primary script from Glottolog\footnote{https://glottolog.org/meta/downloads}, and resource classification (levels 0–5) from \citet{joshi-etal-2020-state}'s taxonomy. To avoid inflated counts from shared datasets appearing across multiple aggregation benchmarks, we deduplicate at the (dataset, language) level while retaining benchmark provenance. Each property is assessed at the dataset level with an associated confidence rating (high, medium, or low) and a supporting evidence passage extracted from the paper. These produced assessments are then surfaced in a web-based human review interface (Figure~\ref{fig:annotate-users}), where reviewers inspect the extracted values and evidence, verify or correct language membership through interactive toggles, and provide final judgments via structured dropdowns. Certain datasets are automatically flagged for mandatory review due to missing or uncertain property values (e.g., unknown native annotator status), indicated by colored \textsc{needs review} badges. 

\subsection{Geographic Distribution}
\label{sec:appendix-geographic}

\begin{figure}[t]
    \centering
    \includegraphics[width=0.85\textwidth]{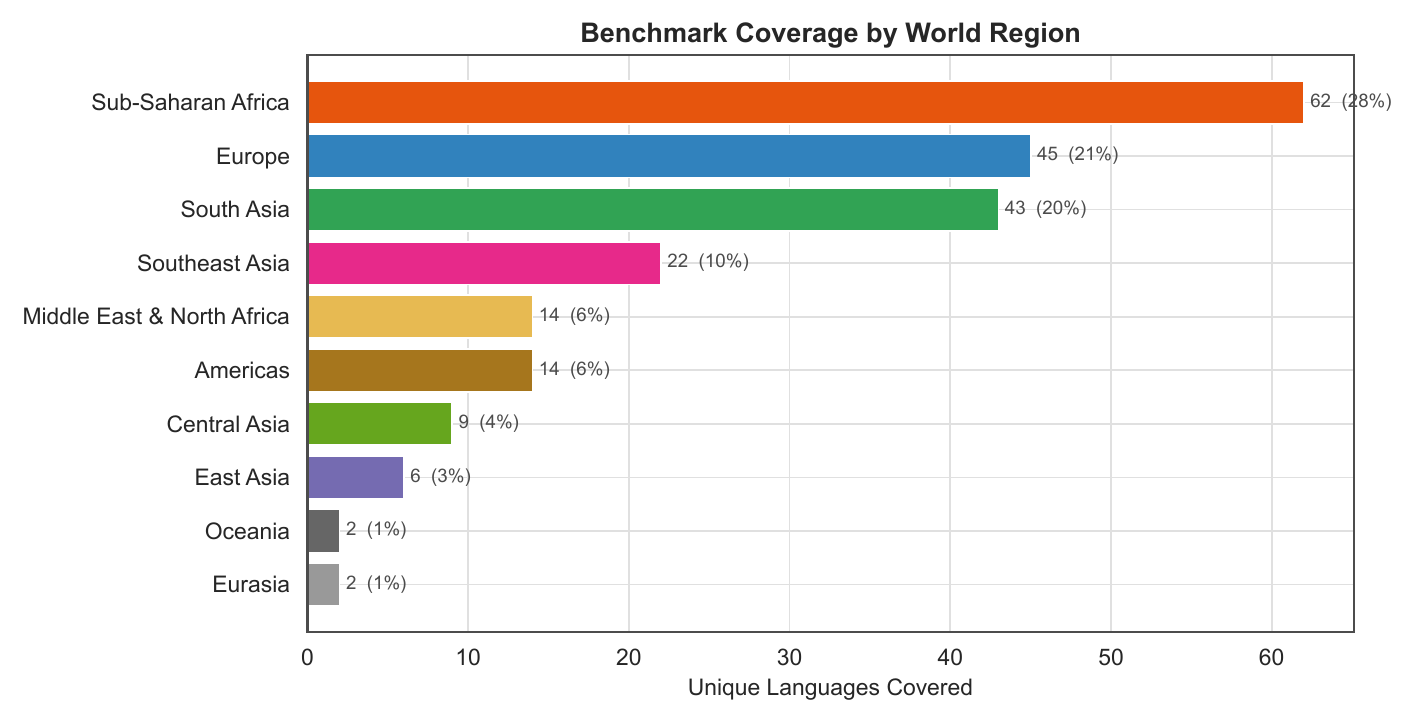}
    \caption{Number of unique languages covered by world region. \textbf{\textit{Takeaway}}: \textit{Coverage is concentrated in Sub-Saharan Africa, Europe, and South Asia; Oceania remains near-zero.}}
    \label{fig:continental-coverage}
\end{figure}

Figure~\ref{fig:continental-coverage} maps languages to macro-regions. Sub-Saharan Africa leads in raw language count (62 languages, 28\%, driven by dedicated datasets), followed by Europe (45, 21\%) and South Asia (43, 20\%). The Americas (14) and Central Asia (9) have gained modest representation through recent datasets, but Oceania remains near-zero (2 languages, 1\%)—a region home to over 1,400 living languages that remain almost entirely outside the evaluation perimeter. 

\begin{figure}[t]
    \centering
    \includegraphics[width=0.65\textwidth]{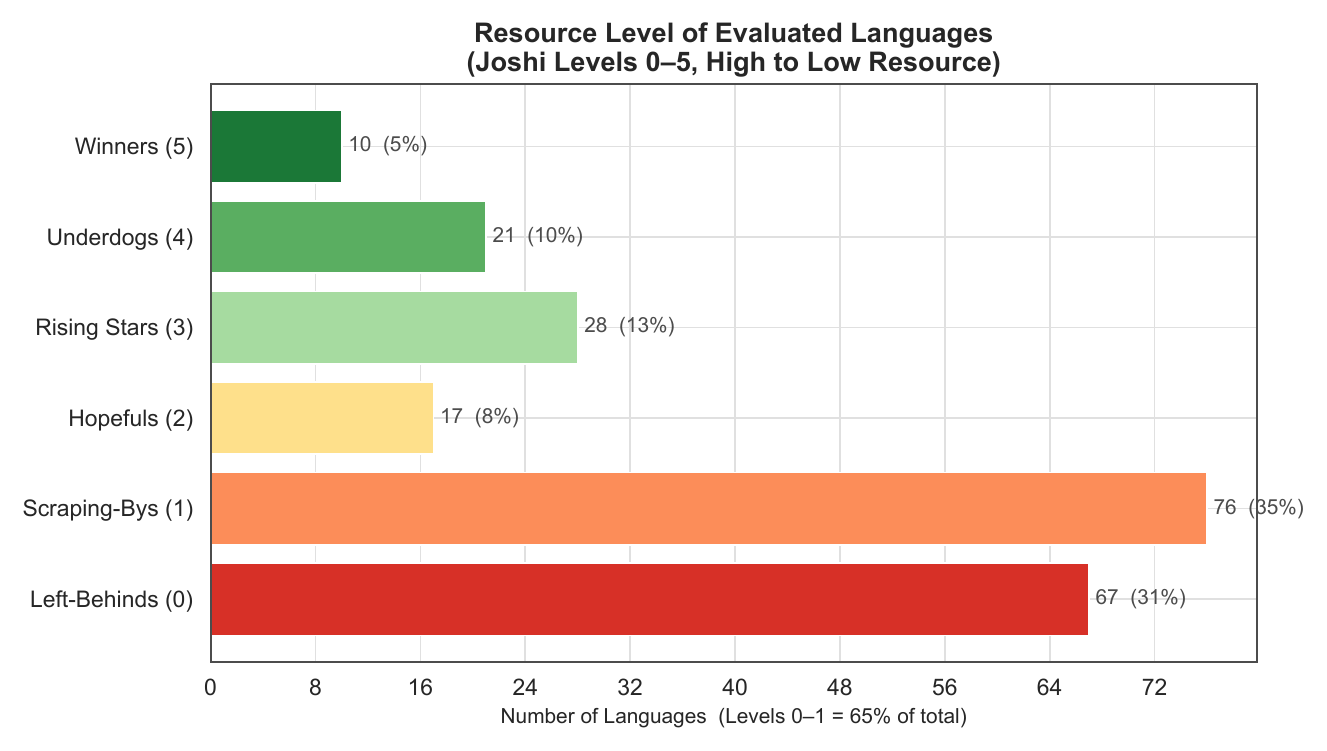}
    \caption{Resource-level distribution of all evaluated languages. \textbf{\textit{Takeaway}}: \textit{65\% of evaluated languages fall in the two lowest resource classes (levels 0--1), yet depth of evaluation remains concentrated in high-resource languages.}}
    \label{fig:resource-pie-appendix}
\end{figure}

\subsection{Resource-Level Distribution}
\label{sec:appendix-resource-level}

Figure~\ref{fig:resource-pie-appendix} shows the resource-level composition of all evaluated languages, complementing the task equity analysis in the section~\ref{sec:representativeness} (Figure~\ref{fig:task-equity-fine}).

\newpage




\begin{figure}[t]
    \centering
    \includegraphics[width=0.7\textwidth]{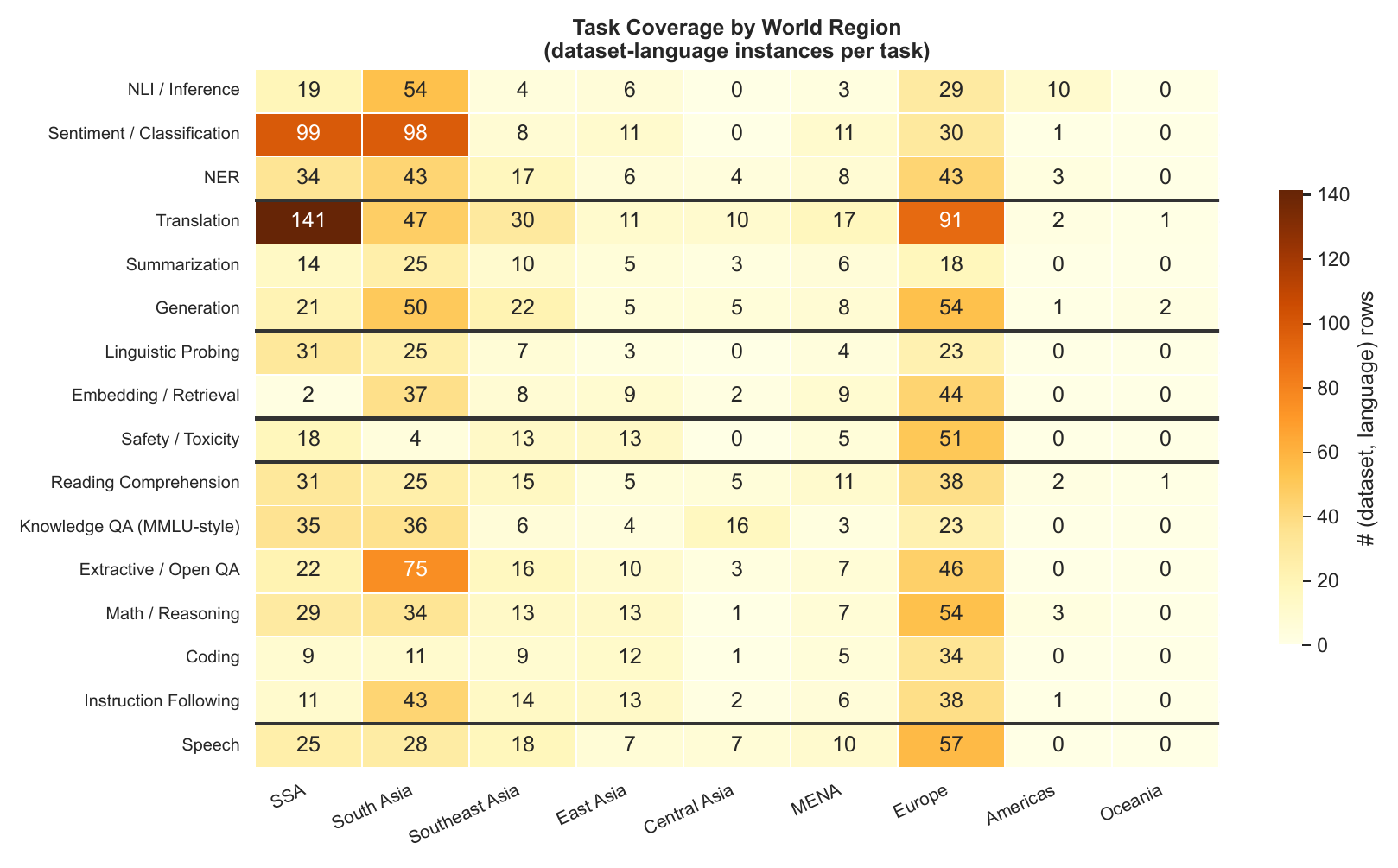}
    \caption{Task category coverage by world region (number of datatset evaluating each task). \textbf{\textit{Takeaway}}: \textit{QA/Knowledge dominates everywhere; specialized tasks (safety, reasoning, coding) are unevenly distributed, with Central Asia and MENA consistently thinnest.}}
    \label{fig:task-region-appendix}
\end{figure}

\subsection{Dataset Source Type Breakdown}
\label{sec:appendix-source-type}

\begin{figure}[t]
    \centering
    \includegraphics[width=0.7\textwidth]{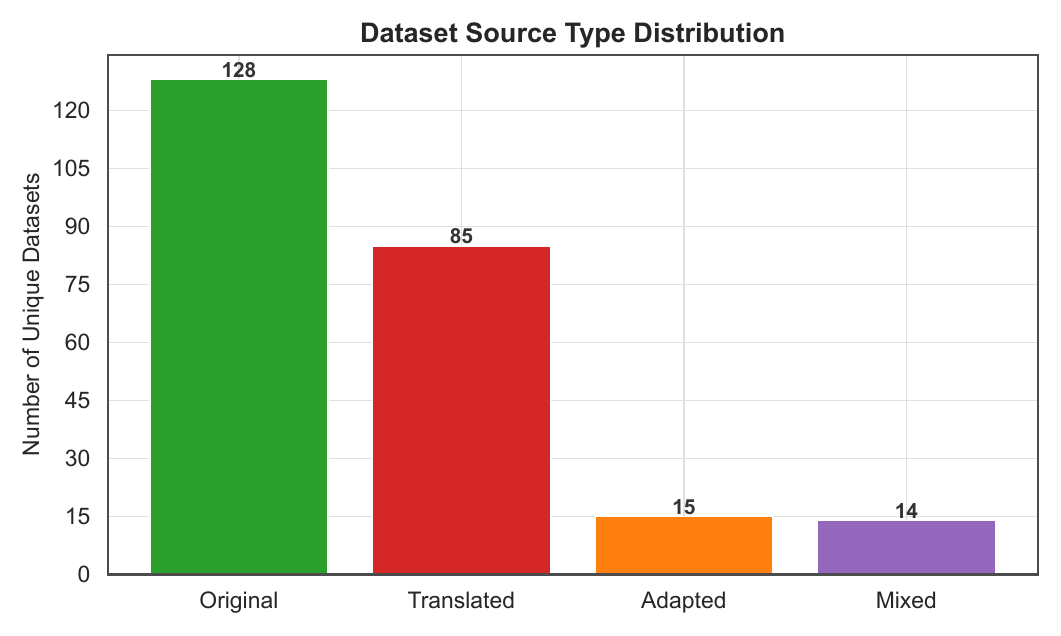}
    \caption{Distribution of dataset construction methods across 242 unique datasets. \textit{35\% of datasets are translated from English; only 53\% are originally authored in the target language(s).}}
    \label{fig:source-type}
\end{figure}

Figure~\ref{fig:source-type} breaks down the 242 unique datasets by construction method. While 128 datasets (53\%) are originally authored, 85 (35\%) are translated from English sources. The remaining datasets are either adapted from existing resources (15, 6\%) or use a mix of methods (14, 6\%). This means that over a third of the evaluation data that multilingual models are tested on originates from English, carrying the associated risks of translationese artifacts and cultural misalignment.

\subsection{Translation Status by Resource Level}
\label{sec:appendix-translation-resource}

\begin{figure}[t]
    \centering
    \includegraphics[width=\textwidth]{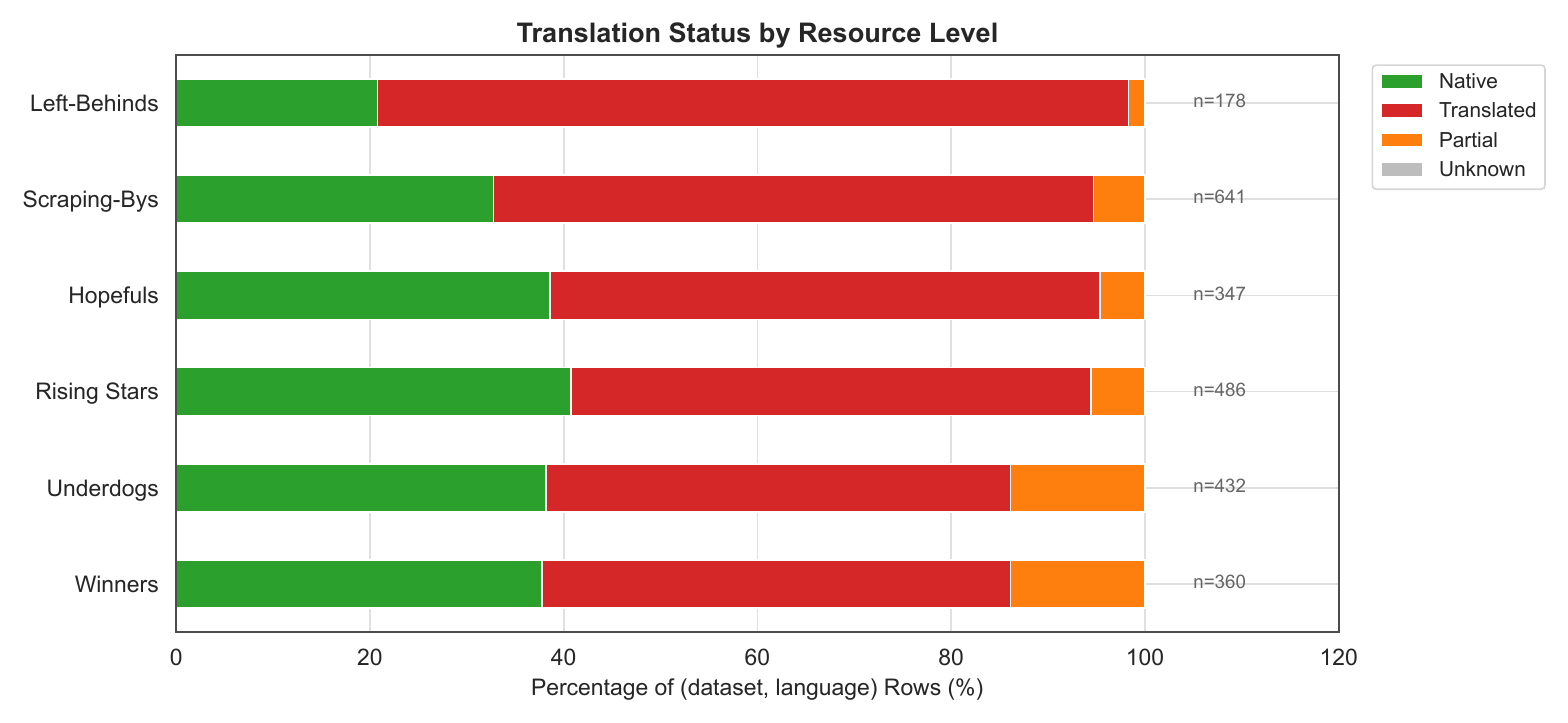}
    \caption{Translation status of dataset--language instances by Joshi resource level. \textbf{\textit{Takeaway}}: \textit{Left-Behinds have the lowest native data share; the most under-resourced languages are predominantly evaluated through translated datasets.}}
    \label{fig:translation-resource}
\end{figure}

Figure~\ref{fig:translation-resource} reveals how translation status interacts with resource level. Across all resource levels, roughly 35--40\% of dataset--language instances involve native data. However, the composition shifts at the extremes: Left-Behinds ($n=178$) have the lowest native share (${\sim}20\%$) and are overwhelmingly evaluated through translated datasets, while Winners ($n=360$) maintain a higher native proportion (${\sim}40\%$) but still rely heavily on translated data. This pattern suggests that the most under-served languages are also the most likely to be evaluated using data that does not originate from their speech communities.

\subsection{Native Annotators by Region}
\label{sec:appendix-native-annotators}

\begin{figure}[h]
    \centering
    \includegraphics[width=\textwidth]{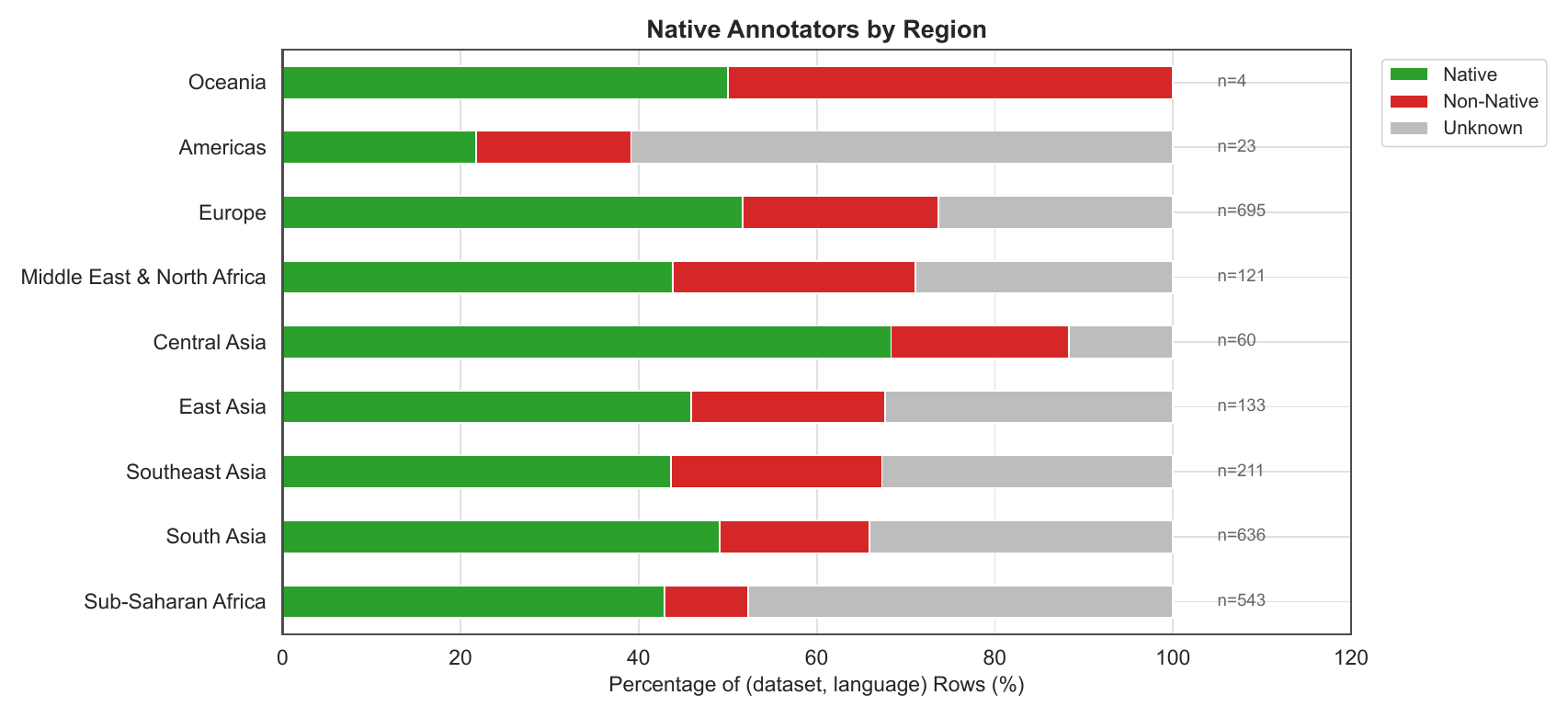}
    \caption{Native vs.\ non-native annotator status by region. \textbf{\textit{Takeaway}}: \textit{Annotator provenance is under-documented; where reported, South Asia and Central Asia lead in native-speaker annotation, while the Americas lag.}}
    \label{fig:native-annotators-region}
\end{figure}

Figure~\ref{fig:native-annotators-region} examines whether dataset annotations were produced by native speakers of the target language, broken down by region. Central Asia and South Asia have the highest native-annotator share (${\sim}50\%$), driven by dedicated regional datasets/benchmarks (e.g., IndicGenBench, INCLUDE). Sub-Saharan Africa ($n=543$) shows ${\sim}45\%$ native annotators but a large ``Unknown'' fraction, reflecting that many aggregation benchmarks do not document annotator provenance. The Americas ($n=23$) have the lowest native share, consistent with minimal community-driven dataset development for indigenous languages.

\newpage
\subsection{Model Release Transparency Audit}
\label{sec:appendix-reporting}

Section~\ref{subsec:reporting-practices} reports summary statistics from our audit of 23 recent model releases. Here we describe the methodology and provide the detailed per-model breakdown.
\begin{figure*}[t]
\centering
\begin{minipage}[t]{0.48\textwidth}
\centering
\includegraphics[width=\textwidth]{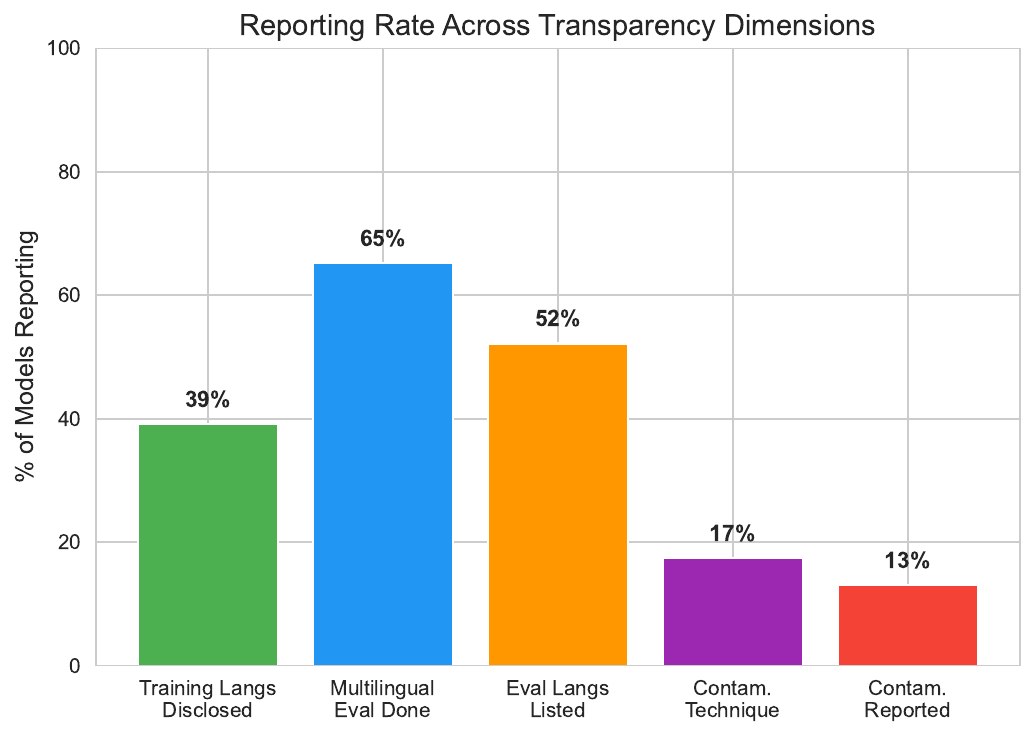}
\end{minipage}%
\hfill
\begin{minipage}[t]{0.48\textwidth}
\centering
\includegraphics[width=\textwidth]{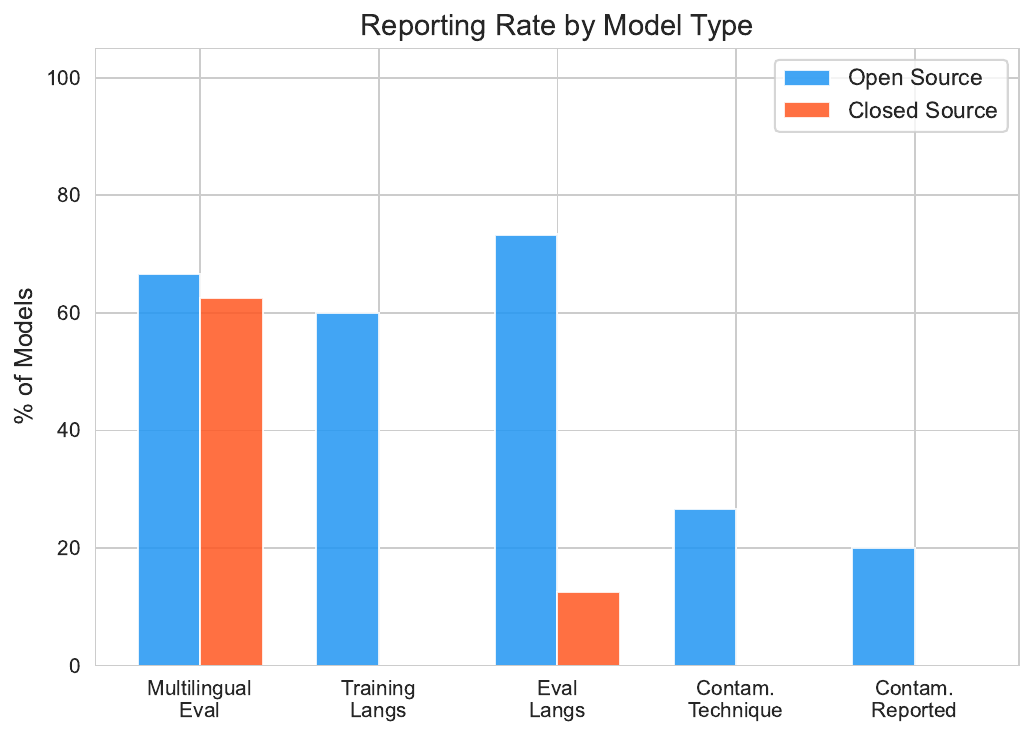}
\end{minipage}
\caption{Multilingual evaluation transparency across 23 recent model releases (15 open-weight, 8 closed). \textbf{Left:} Reporting rate across five transparency dimensions---only 65\% report any multilingual evaluation, and just 13\% report contamination results. \textbf{Right:} Breakdown by model type---closed models disclose significantly less across all dimensions.}
\label{fig:reporting-transparency}
\end{figure*}

\paragraph{Methodology}
We examine technical reports, model cards, and blog posts for 23 model releases spanning 2024--2026 (15 open-weight, 8 closed-models). Each release is scored on five binary transparency dimensions: (1)~whether training language composition is explicitly disclosed, (2)~whether any multilingual evaluation datasets is reported, (3)~whether the specific evaluation languages are listed, (4)~whether a contamination detection technique is named, and (5)~whether contamination results are reported. The composite transparency score is the sum (0--5).

\paragraph{Detailed Findings}


\paragraph{Models with sparse multilingual evaluation evidence} include GPT, Grok, Mistral~3, BharatGen, Command-R+, OLMo, and Kimi~K2.5. Kimi~K2.5, for example, is described by its authors as ``very English centric.'' Similarly, while Sarvam is trained on 22 Indian languages, its reported evaluation appears to center on a newly introduced dataset/benchmark scored using an LLM judge, without corresponding results on other widely used multilingual datasets. Consequently, the multilingual capabilities of these releases remain difficult to verify comprehensively, given the limited publicly reported evaluation evidence.

\paragraph{Contamination testing} is reported by only 3 models: Llama-3 (ConTAM), Gemma (quality reweighting), and GPT-OSS (TroubleshootingBench). All are open-weight. Even these three use techniques designed for English benchmarks; cross-lingual contamination pathways like translated dataset leakage, parallel corpus overlap, instruction-tuning propagation go unexamined.

\paragraph{Datasets concentration:} INCLUDE, FLORES-200, and MGSM are the three most frequently cited multilingual datasets, each appearing in only 4 of 23 reports. Natively authored, culturally grounded datasets like AfroBench, IrokoBench, MILU, CulturalBench, TUMLU are almost never cited.

\paragraph{Open weight vs.\ closed:} Open-weight models average 2.47/5 on the transparency score vs.\ 0.75/5 for closed models. The gap is largest on training language disclosure (60\% vs.\ 0\%), evaluation language listing (73\% vs.\ 12\%), and contamination reporting (20\% vs.\ 0\%). The Aya model family (Aya-23, Aya-Expanse, Tiny~Aya) stands out as the transparency exemplar, reporting multilingual datasets across 23--70 languages. However, even the most transparent open-weight models rarely report contamination testing for multilingual datasets.

\paragraph{Multilingual safety evaluation} is reported by only 4 of 23 models: Tiny~Aya, Claude~Opus~4.6, Claude~Sonnet~4.6, and CommandA. All other models either test safety only in English or do not report safety testing at all.

\newpage
\section{Contamination Detection Results}
\label{sec:appendix-contamination}

\begin{table}[!htbp]
\centering
\resizebox{\textwidth}{!}{%
\begin{tabular}{l | cccccccccc}
\toprule
& \textsc{Llama-3.1-8B} & \textsc{Llama-3.1-8B-it} & \textsc{Mistral-7B-v0.3} & \textsc{Mistral-7B-v0.3-it} & \textsc{Gemma-2-9B-it} & \textsc{Gemma-2-9B} & \textsc{Aya-23-8B} & \textsc{Gemma-7B-it} & \textsc{Llama-2-7B-it} & \textsc{Mistral-7B-v0.1-it} \\
\midrule
\textsc{Flores}      & \cont & \cont & \cont & \cont & \cont & \cont & \cont & --        & --        & -- \\
\textsc{Paws-X}      & \cont & \cont & \cont & \cont & \cont & \cont & \notcont  & \cont  & \cont     & \cont \\
\textsc{XCopa}       & \cont & \cont & \cont & \cont & \cont & \cont & \notcont  & \cont  & \cont     & \cont \\
\textsc{XLSum}       & \notcont & \notcont & \cont & \cont & \cont & \cont & \cont & --     & --        & -- \\
\textsc{Xnli}        & \cont & \cont & \cont & \cont & \cont & \cont & \cont & \notcont  & \notcont  & \notcont \\
\textsc{XQuad}       & \cont & \cont & \cont & \cont & \cont & \cont & \cont & \notcont  & \cont     & \cont \\
\textsc{XStoryCloze} & \cont & \cont & \cont & \cont & \cont & \cont & \cont & \notcont  & \notcont  & \notcont \\
\bottomrule
\end{tabular}%
}
\caption{Benchmark contamination presence across all evaluated models based on \cite{ahuja-etal-2024-megaverse}, \cite{ahuja2024contaminationreportmultilingualbenchmarks}, and \cite{yao-etal-2024-data}. \cont\ = \textbf{contaminated}, \notcont\ = \textbf{not contaminated}, -- = not evaluated.}
\label{tab:contamination_combined}
\end{table}

\begin{table}[!htbp]
\centering
\resizebox{\textwidth}{!}{%
\begin{tabular}{l cccccccccccc}
\toprule
& \textsc{Phi2} & \textsc{Phi3} & \textsc{Phi3.5-mini} & \textsc{Phi3.5-MoE} & \textsc{GrinMoE} & \textsc{Abel-7B} & \textsc{LLaMA2} & \textsc{Mistral} & \textsc{Qwen2} & \textsc{GLM4} & \textsc{LLaMA3} & \textsc{Reflection} \\
& 2.7B & 3.8B & 3.8B & 3.8B$\times$16 & 3.8B$\times$16 & 7B & 7B & 7B & 7B & 9B & 70B & 70B \\
\midrule
MMLU           & 23.83 & 67.27 & 68.64 & 76.62 & 77.55 & 47.08 & 44.88 & 57.29 & 69.05 & 67.36 & 78.55 & 75.83 \\
MMLU-g         & 25.02 & 85.29 & 87.00 & 91.65 & 92.83 & 68.37 & 72.87 & 82.71 & 89.23 & 84.91 & 92.17 & 88.37 \\
\textit{difference} & 1.20 & 18.02 & 18.36 & 15.03 & 15.28 & \textbf{21.29} & \textbf{27.99} & \textbf{25.42} & 20.18 & 17.55 & 13.62 & 12.54 \\
\midrule
ARC-C          & 42.92 & 80.20 & 59.56 & 54.78 & 63.57 & 50.34 & 36.18 & 64.08 & 84.81 & 86.35 & 61.52 & 56.74 \\
ARC-C-g        & 47.27 & 92.15 & 93.94 & 96.50 & 96.25 & 66.04 & 44.71 & 85.75 & 95.22 & 91.81 & 95.99 & 94.45 \\
\textit{difference} & \textbf{4.35} & 11.95 & \textbf{34.38} & \textbf{41.72} & 32.68 & 15.70 & 8.53 & 21.67 & 10.41 & \textbf{5.46} & \textbf{34.47} & \textbf{37.71} \\
\midrule
MathQA         & 31.32 & 41.14 & 41.14 & 37.42 & 47.67 & 34.30 & 28.71 & 36.88 & 44.36 & 43.05 & 56.52 & 58.29 \\
MathQA-g       & 38.70 & 49.06 & 47.38 & 43.96 & 56.27 & 35.71 & 36.18 & 45.77 & 49.03 & 56.04 & 61.84 & 63.92 \\
\textit{difference} & 7.38 & 7.92 & 6.24 & 6.54 & 8.60 & \textbf{1.41} & 7.47 & 8.89 & 4.67 & 12.99 & 5.32 & 5.63 \\
\bottomrule
\end{tabular}%
}
\caption{Detecting inadvertent contamination in popular open-source LLMs from \citet{yao-etal-2024-data}. Each benchmark is tested in its original form and a paraphrased variant (-g); the \textit{difference} row shows the generalizability gap. \textbf{Bold} values indicate significantly lower generalizability, implying potential contamination.}
\label{tab:contamination_crosslingual}
\end{table}

\begin{table}[!htbp]
\centering
\resizebox{\linewidth}{!}{%
\begin{tabular}{lcccccccccc}
\toprule
Model & PAW-SX & Global MMLU & IndicParam & MILU & MMMLU & INCLUDE & FLORES & MGSM & AFRiMGSM & Overall \\
\midrule
  Aya-Expanse-32B & \textbf{100\% (4/4)} & 0\% (0/42) & 0\% (0/6) & 0\% (0/11) & 0\% (0/14) & 0\% (0/44) & 4\% (1/28) & \textbf{55\% (6/11)} & \textbf{16\% (3/19)} & 7.8\% \\
  Gemma-3-12B & \textbf{50\% (2/4)} & 12\% (5/40) & 0\% (0/7) & 0\% (0/11) & \textbf{14\% (2/14)} & \textbf{5\% (2/44)} & 0\% (1/204) & \textbf{55\% (6/11)} & \textbf{16\% (3/19)} & 5.9\% \\
  Tiny-Aya-Global & \textbf{100\% (4/4)} & 0\% (0/42) & 0\% (0/12) & 0\% (0/11) & 0\% (0/14) & 0\% (0/44) & 2\% (5/204) & \textbf{55\% (6/11)} & \textbf{26\% (5/19)} & 5.5\% \\
  Aya-Expanse-8B & \textbf{75\% (3/4)} & 0\% (0/42) & 10\% (1/10) & 0\% (0/11) & 0\% (0/14) & 0\% (0/44) & 1\% (2/204) & \textbf{64\% (7/11)} & \textbf{26\% (5/19)} & 5.0\% \\
  Qwen3-4B & \textbf{100\% (4/4)} & 0\% (0/42) & 8\% (1/12) & 0\% (0/11) & 0\% (0/14) & 0\% (0/44) & 0\% (1/204) & \textbf{36\% (4/11)} & \textbf{21\% (4/19)} & 3.9\% \\
  Phi-4 & \textbf{25\% (1/4)} & 0\% (0/42) & \textbf{8\% (1/12)} & 0\% (0/11) & 0\% (0/14) & 2\% (1/44) & 1\% (2/204) & \textbf{18\% (2/11)} & \textbf{21\% (4/19)} & 3.0\% \\
  Llama-3.1-8B & \textbf{25\% (1/4)} & 0\% (0/42) & 0\% (0/10) & 0\% (0/11) & 0\% (0/14) & 0\% (0/44) & 1\% (3/204) & \textbf{36\% (4/11)} & \textbf{20\% (2/10)} & 2.9\% \\
\midrule
  Overall & 67.9\% & 1.7\% & 4.3\% & 0.0\% & 2.0\% & 1.0\% & 1.2\% & 45.5\% & 21.0\% & 4.6\% \\
\bottomrule
\end{tabular}
}
\caption{Contamination rate per model and dataset. Each cell shows the fraction of
languages flagged as contaminated ($p < 0.05$), with the count in parentheses.
\textbf{Bold} indicates the dataset is contaminated as a whole
(Fisher's combined $p < 0.05$).}
\label{tab:contamination_results}
\end{table}

\subsection{Data Contamination Quiz Results}
\label{sec:appendix-dcq}

We apply the Data Contamination Quiz (DCQ)~\cite{} to 85 dataset--language partitions, with 100 instances per partition. Table~\ref{tab:dcq_results} reports the maximum recognition rate, averaged over languages within each dataset. Table~\ref{tab:dcq_language_groups} compares English partitions with the mean over non-English partitions where both are available. Because each quiz has five options, 20\% is the random-chance floor. These are recognition rates rather than estimates of the fraction of training data contaminated.

Eight partitions have perturbation distance below 10\%, making their quizzes close to unsolvable and their low scores uninformative. The bias-detection abstention rate is also near zero for most models, indicating that models often guess rather than select ``none of these.'' A byte-identical replicate of the English GSM8K items under the AfriMGSM and MGSM dataset names produces a mean absolute difference of 9.7 percentage points and a maximum difference of 23 points across models. Individual DCQ values should therefore be treated as approximate.

\begin{table}[!htbp]
\centering
\scriptsize
\resizebox{\linewidth}{!}{%
\begin{tabular}{lcccccccc}
\toprule
\textbf{Model} & \textbf{AfriMGSM} & \textbf{Global MMLU} & \textbf{INCLUDE} & \textbf{IndicParam} & \textbf{MGSM} & \textbf{MILU} & \textbf{MMMLU} & \textbf{PAWS-X} \\
\midrule
Aya-Expanse-32B & 21.3 & 28.7 & 29.7 & 16.2 & 47.4 & 27.0 & 31.8 & 54.4 \\
Aya-Expanse-8B & 10.8 & 12.6 & 11.5 & 8.7 & 22.3 & 13.6 & 12.4 & 38.9 \\
Tiny-Aya-Global & 27.5 & 21.0 & 19.1 & 15.1 & 34.6 & 16.7 & 22.8 & 11.4 \\
Gemma-3-12B & 39.4 & 21.5 & 40.6 & 36.0 & 56.6 & 36.6 & 19.7 & 41.3 \\
Llama-3.1-8B & 9.2 & 8.7 & 11.1 & 9.9 & 17.9 & 9.8 & 7.3 & 33.2 \\
Phi-4 & 2.3 & 22.6 & 10.6 & 13.2 & 10.6 & 24.6 & 36.8 & 20.6 \\
Qwen3-4B & 18.1 & 7.7 & 9.3 & 3.8 & 26.8 & 13.7 & 6.5 & 28.2 \\
\midrule
Mean & 18.4 & 17.5 & 18.8 & 14.7 & 30.9 & 21.7 & 19.6 & 32.6 \\
\bottomrule
\end{tabular}%
}
\caption{DCQ maximum recognition rate (\%) by model and benchmark, averaged over each benchmark's language partitions. The random-chance floor is 20\%. The maximum rate selects the best raw hit rate across the tested answer positions and is therefore an optimistic estimate.}
\label{tab:dcq_results}
\end{table}

\begin{table}[!htbp]
\centering
\small
\begin{tabular}{lcc}
\toprule
\textbf{Benchmark} & \textbf{English} & \textbf{Non-English Mean} \\
\midrule
AfriMGSM & 45.7\% & 15.3\% \\
Global MMLU & 26.1\% & 16.6\% \\
MGSM & 44.0\% & 29.6\% \\
MILU & 37.9\% & 18.6\% \\
PAWS-X & 47.6\% & 30.1\% \\
\bottomrule
\end{tabular}
\caption{DCQ maximum recognition rate averaged across the seven models for English and non-English language partitions. INCLUDE, IndicParam, and MMMLU are omitted because the evaluated partitions do not include English.}
\label{tab:dcq_language_groups}
\end{table}

\begin{table}[!htbp]
\centering
\small
\begin{tabular}{llc}
\toprule
\textbf{Type} & \textbf{Name} & \textbf{Release Date} \\
\midrule
\multirow{7}{*}{Model}
  & Aya-Expanse-32B   & December 2024\\
  & Aya-Expanse-8B    & December 2024\\
  & Tiny-Aya-Global   & February 2026\\
  & Gemma-3-12B       & March 2025\\
  & Llama-3.1-8B      & July 2024\\
  & Phi-4             & December 2024\\
  & Qwen3-4B          & April 2025\\
\midrule
\multirow{9}{*}{Datasets}
  & PAW-SX      & November 2019\\
  & Global MMLU & December 2024\\
  & IndicParam  & November 2025\\
  & MILU        & November 2024\\
  & MMMLU       & September 2020\\
  & INCLUDE     & November 2024\\
  & FLORES      & June 2021\\
  & MGSM        & October 2022\\
  & AFRiMGSM    & May 2024\\
\bottomrule
\end{tabular}
\caption{Release dates of models and datasets/benchmarks used in this study.}
\label{tab:release_dates}
\end{table}

\section{Released Resources and Interactive Dashboard}
\label{sec:appendix-dashboard}

To facilitate community engagement with these findings, we will release three resources upon acceptance. First, we will provide the \textbf{structured metadata annotations} for all 51 benchmarks and 242 datasets, including per-language fields for script, language family, resource level, task categories, translation status, native annotator provenance, and cultural grounding. Second, we will release the \textbf{full analysis codebase} used to produce all figures, tables, and statistics in this paper, enabling researchers to reproduce our analysis and extend it as new datasets appear. Third, we provide an \textbf{interactive web dashboard} that allows users to explore coverage and representativeness statistics, browse individual datasets via a searchable explorer, and look up per-language dataset coverage. The dashboard is designed to be a living resource that will be updated as new datasets are added to the survey.

\begin{figure}[h]
    \centering
    \includegraphics[width=\textwidth]{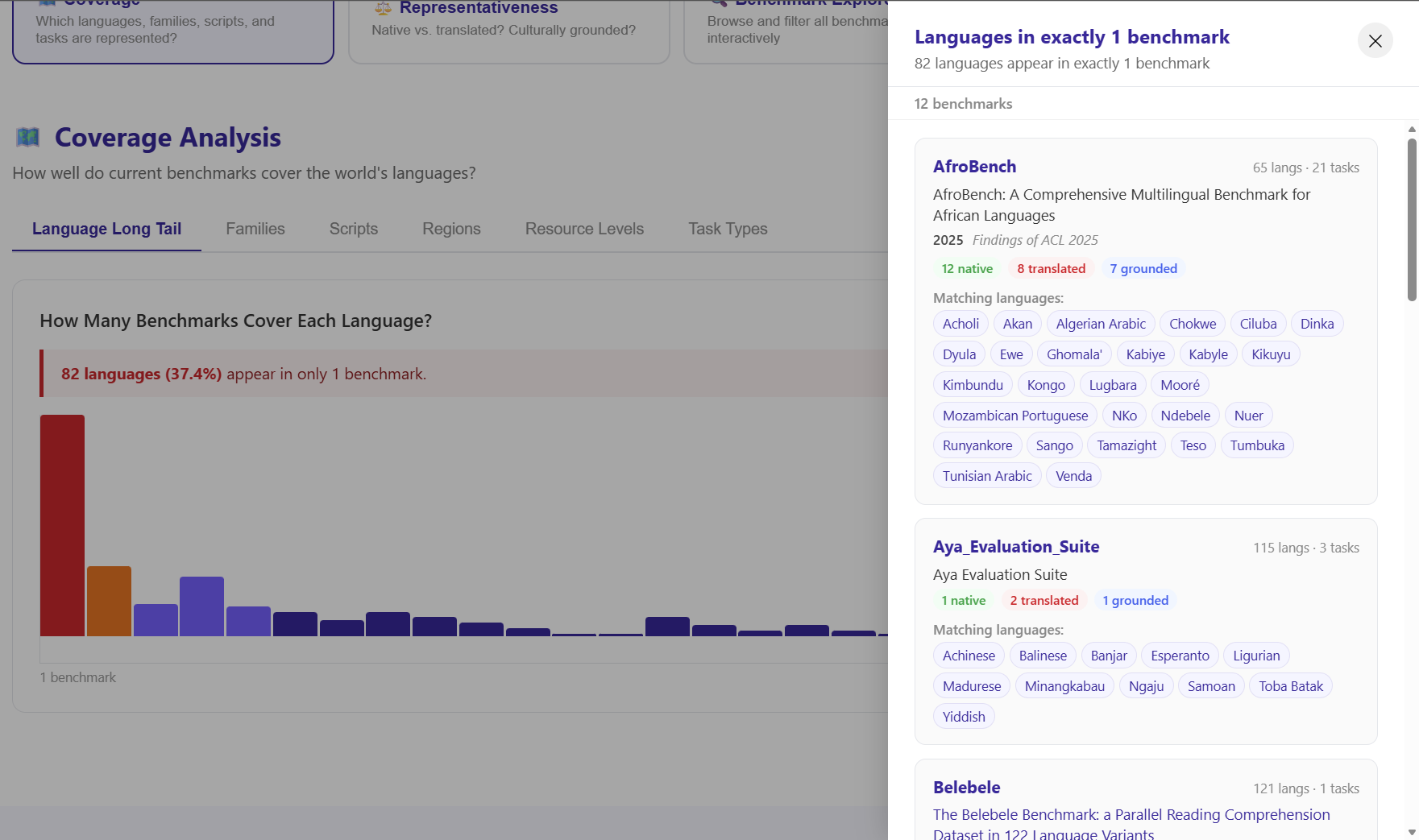}
    \caption{Coverage Analysis view with interactive drill-down. Clicking on a bar in the language long-tail chart opens a panel listing the specific benchmarks that cover languages appearing in exactly that number of benchmarks, along with matching languages for each benchmark.}
    \label{fig:dashboard-coverage}
\end{figure}

\begin{figure}[h]
    \centering
    \includegraphics[width=\textwidth]{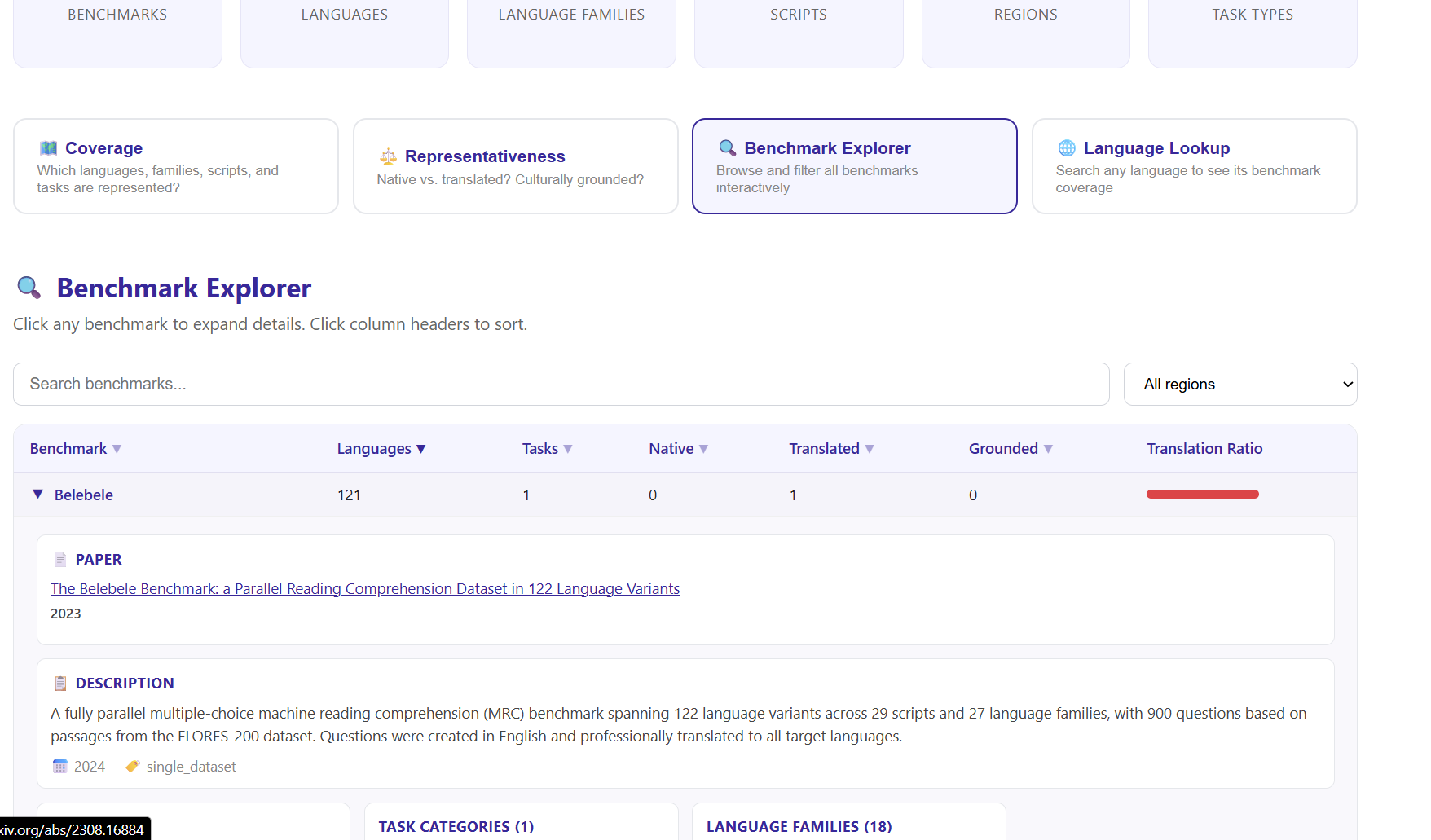}
    \caption{Benchmark Explorer view. Users can search, sort by column, and filter by region. Expanding a row reveals the paper reference, description, year, benchmark type, task categories, and full language family breakdown.}
    \label{fig:dashboard-explorer}
\end{figure}

The dashboard provides four main views:

\begin{itemize}
    \item \textbf{Coverage Analysis}: Interactive visualisations of language, family, script, regional, resource-level, and task-type distributions, mirroring and extending the figures in this paper. Users can click on individual bars to drill down into which datasets cover specific languages (Figure~\ref{fig:dashboard-coverage}).
    \item \textbf{Representativeness}: Translation status and cultural grounding breakdowns by region and resource level.
    \item \textbf{Benchmark Explorer}: A searchable, sortable table of all 51 benchmarks with expandable rows showing paper links, descriptions, language lists, task categories, and language family coverage (Figure~\ref{fig:dashboard-explorer}).
    \item \textbf{Language Lookup}: A per-language search interface that returns all datasets/benchmarks covering a given language, with metadata on translation status and task types.
\end{itemize}


\begin{figure}[h]
    \centering
    \includegraphics[width=\textwidth]{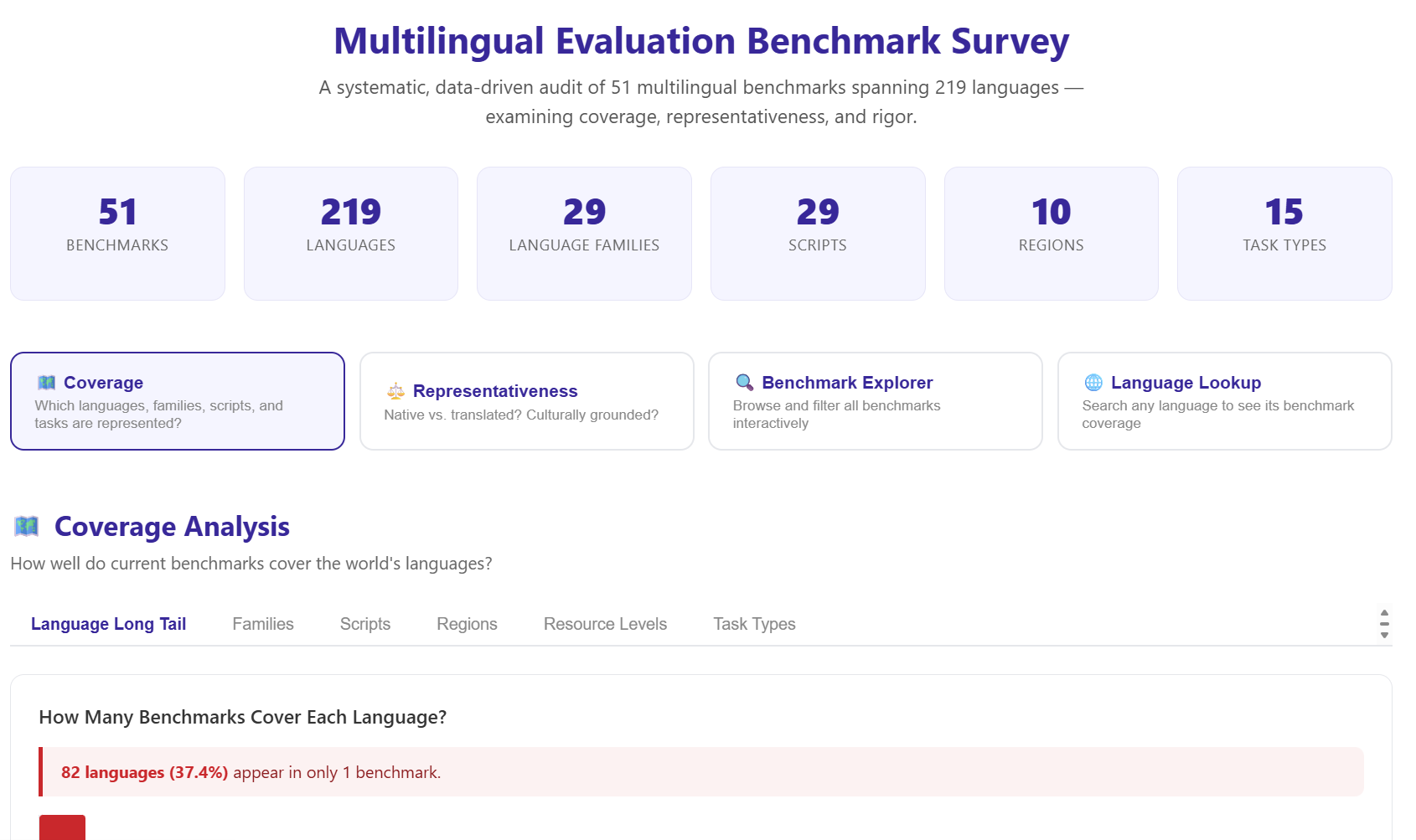}
    \caption{Landing page of the interactive dashboard showing summary statistics (51 benchmarks, 219 languages, 29 families, 29 scripts, 10 regions, 15 task types) and navigation to the four main views: Coverage, Representativeness, Benchmark Explorer, and Language Lookup.}
    \label{fig:dashboard-landing}
\end{figure}

\begin{table*}[t]
\centering
\resizebox{\textwidth}{!}{%
\begin{tabular}{@{}r l l l l l@{}}
\toprule
\textbf{\#} & \textbf{Benchmark} & \textbf{Task Categories} & \textbf{Year} & \textbf{Venue} & \textbf{Citation} \\
\midrule
  1 & MMTEB & Classification / QA / Knowledge (550 tasks)  & 2025 & ICLR & \citep{enevoldsen2025mmteb} \\
  2 & Belebele & QA / Knowledge & 2024 & ACL & \citep{bandarkar2024belebele} \\
  3 & Aya\_Evaluation\_Suite & Generation & 2024 & ACL & \citep{singh2024aya} \\
  4 & FLORES-101 & Translation & 2022 & TACL & \citep{goyal2022flores} \\
  5 & XTREME\_UP & Linguistic Probing, Other, QA / Knowledge, \dots & 2023 & ACL & \citep{ruder2023xtreme} \\
  6 & MEGA & NER, NLI / Inference, QA / Knowledge, \dots & 2023 & EMNLP & \citep{ahuja-etal-2023-mega} \\
  7 & AfroBench & Linguistic Probing, Math / Reasoning, NER, \dots & 2025 & ACL & \citep{ojo-etal-2025-afrobench} \\
  8 & BUFFET & NER, NLI / Inference, QA / Knowledge, \dots & 2024 & NAACL & \citep{asai2024buffet} \\
  9 & XTREME-R & Embedding / Retrieval, Linguistic Probing, \dots & 2021 & EMNLP & \citep{ruder-etal-2021-xtreme} \\
  10 & XTREME & Embedding / Retrieval, NLI / Inference, \dots & 2020 & ICML & \citep{siddhant2020xtreme} \\
  11 & MUG\_Eval & Coding, Instruction Following, Math / Reasoning & 2025 & EMNLP & \citep{song-etal-2025-mug-eval} \\
  12 & MMLU-ProX & QA / Knowledge & 2025 & EMNLP & \citep{xuan2025mmlu} \\
  13 & MaXIFE & Instruction Following & 2025 & ACL & \citep{liu-etal-2025-maxife} \\
  14 & IndiSentiment140 & Sentiment / Classification & 2024 & NAACL & \citep{kumar2024indisentiment140} \\
  15 & JailNewsBench & Safety / Toxicity & 2026 & ICLR & \citep{kanekojailnewsbench} \\
  16 & IndicXTREME & Embedding / Retrieval, NER, NLI / Inference, \dots & 2023 & ACL & \citep{doddapaneni2023towards} \\
  17 & IrokoBench & Math / Reasoning, NLI / Inference, QA / Knowledge & 2025 & NAACL & \citep{adelani-etal-2025-irokobench} \\
  18 & XGLUE & Embedding / Retrieval, Linguistic Probing, \dots & 2020 & EMNLP & \citep{liang-etal-2020-xglue} \\
  19 & BenchMAX & Coding, Instruction Following, \dots & 2025 & EMNLP & \citep{huang-etal-2025-benchmax} \\
  20 & PTP & Safety / Toxicity & 2024 & COLM & \citep{jainpolyglotoxicityprompts} \\
  21 & BeMyCheese & Sentiment / Classification & 2026 & ArXiv & \citep{van2026my} \\
  22 & XNLI & NLI / Inference & 2018 & EMNLP & \citep{conneau-etal-2018-xnli} \\
  23 & IndicIFEval & Instruction Following & 2026 & ArXiv & \citep{jayakumar2026indicifeval} \\
  24 & INDIC-DIALECT & QA / Knowledge, Sentiment / Classification, \dots & 2026 & ArXiv & \citep{sharma2026indic} \\
  25 & LinguaSafe & Safety / Toxicity & 2025 & ArXiv & \citep{ning2025linguasafe} \\
  26 & MAKIEVAL & Sentiment / Classification & 2025 & EMNLP & \citep{zhao-etal-2025-makieval} \\
  27 & GSM8K-Indic & Math / Reasoning & 2025 & --- & sarvamai/gsm8k-indic \\
  28 & IndicNLG & Generation, NLI / Inference, QA / Knowledge, \dots & 2022 & EMNLP & \citep{kumar-etal-2022-indicnlg} \\
  29 & IndicNLPSuite & Embedding / Retrieval, NER, NLI / Inference, \dots & 2020 & EMNLP & \citep{kakwani2020indicnlpsuite} \\
  30 & IndicParam & QA / Knowledge & 2025 & ArXiv & \citep{maheshwari2025indicparam} \\
  31 & MILU & QA / Knowledge & 2025 & NAACL & \citep{verma-etal-2025-milu} \\
  32 & Naamapadam & NER & 2023 & ACL & \citep{mhaske-etal-2023-naamapadam} \\
  33 & TriviaQA-Indic & QA / Knowledge & 2025 & --- & sarvamai/trivia-qa-indic-mcq \\
  34 & XQuAD & QA / Knowledge & 2020 & ACL & \citep{artetxe-etal-2020-cross} \\
  35 & BoolQ-Indic & QA / Knowledge & 2025 & --- & sarvamai/boolq-indic \\
  36 & MMLU-Indic & QA / Knowledge & 2025 & --- & sarvamai/mmlu-indic \\
  37 & MultiJail & Safety / Toxicity & 2024 & ICLR 2024 & \citep{deng2024multilingual} \\
  38 & IndicMMLU-Pro & QA / Knowledge & 2025 & ArXiv & \citep{kj2025indicmmluprobenchmarkingindiclarge} \\
  39 & M3Exam & QA / Knowledge & 2023 & NeurIPS 2023 & \citep{zhang2023m3exam} \\
  40 & TUMLU & QA / Knowledge & 2025 & ACL & \citep{isbarov-etal-2025-tumlu} \\
  41 & XORQA & Embedding / Retrieval & 2021 & NAACL & \citep{asai-etal-2021-xor} \\
  42 & CL-IFEval & Instruction Following, Math / Reasoning & 2025 & ArXiv & \citep{ojewale2026multilingualfunctionalevaluationlarge} \\
  43 & Uhura & QA / Knowledge & 2024 & ArXiv & \citep{bayes2024uhurabenchmarkevaluatingscientific} \\
  44 & IberoBench & Linguistic Probing, Math / Reasoning, \dots & 2025 & COLING 2025 & \citep{baucells-etal-2025-iberobench} \\
  45 & SeaHELM & Generation, Instruction Following, \dots & 2025 & Findings of ACL 2025 & \citep{susanto-etal-2025-sea} \\
  46 & BHASA & Linguistic Probing, Math / Reasoning, NER, \dots & 2023 & ArXiv & \citep{leong2023bhasa} \\
  47 & M-IFEval & Instruction Following & 2025 & Findings of NAACL 2025 & \citep{dussolle-etal-2025-ifeval} \\
  48 & MUCH & Sentiment / Classification & 2026 & LREC & \citep{dentan2025much} \\
  49 & FluidQA & QA / Knowledge, Sentiment / Classification & 2025 & EMNLP 2025 & \citep{park-etal-2025-fluid} \\
  50 & BhashaBench\_V1 & QA / Knowledge & 2025 & ArXiv & \citep{devane2025bhashabenchv1comprehensivebenchmark} \\
  51 & CulturalBench & QA / Knowledge & 2025 & ACL 2025 & \citep{chiu-etal-2025-culturalbench} \\
\bottomrule
\end{tabular}%
}
\caption{Comprehensive overview of all 51 multilingual evaluation benchmarks surveyed.}
\label{tab:appendix-benchmarks}
\end{table*}

\end{document}